\documentclass[runningheads]{llncs}

\usepackage[T1]{fontenc}

\usepackage{times}
\usepackage{epsfig}
\usepackage{amsmath}
\usepackage{amssymb}
\usepackage{booktabs}
\usepackage[accsupp]{axessibility}
\usepackage{subcaption}
\usepackage{adjustbox}

\usepackage{natbib}
\usepackage{graphicx}
\usepackage{txfonts}
\usepackage{hyperref}

\usepackage{color}
\usepackage{float}
\usepackage[rflt]{floatflt} 
\usepackage{mathtools}
\usepackage[normalem]{ulem}
\usepackage{float} 
\usepackage[table]{xcolor} 

\newcommand{\off}[2]{{}{}}

\usepackage[ruled,vlined,lined,linesnumbered]{algorithm2e}
\usepackage{dsfont}

\usepackage[capitalize]{cleveref}
\crefname{section}{Sec.}{Secs.}
\Crefname{section}{Section}{Sections}
\Crefname{table}{Table}{Tables}
\crefname{table}{Tab.}{Tabs.}

\begin{document}
\title{Automatic Wood Pith Detector: Local Orientation Estimation and Robust Accumulation}
\titlerunning{Automatic Wood Pith Detector}
%
\author{Henry Marichal\inst{1} \and
Diego Passarella\inst{2}\and
Gregory Randall\inst{1}}
\authorrunning{Marichal et al.}
%
\institute{Instituto de Ingeniería Eléctrica, Facultad de Ingeniería, Universidad de la República, Montevideo, Uruguay \and
Centro Universitario Regional Noreste, Universidad de la República, Tacuarembo, Uruguay}
\maketitle              
\begin{abstract}
A fully automated technique for wood pith detection (APD), relying on the concentric shape of the structure of wood ring slices, is introduced. The method estimates the ring's local orientations using the 2D structure tensor and finds the pith position, optimizing a cost function designed for this problem. 
We also present a variant (APD-PCL), using the parallel coordinates space, that enhances the method's effectiveness when there are no clear tree ring patterns. Furthermore, refining previous work by Kurdthongmee, a YoloV8 net is trained for pith detection, producing a deep learning-based approach to the same problem (APD-DL). 
All methods were tested on seven datasets, including images captured under diverse conditions (controlled laboratory settings, sawmill, and forest) and featuring various tree species (\textit{Pinus taeda, Douglas fir, Abies alba}, and \textit{Gleditsia triacanthos}). All proposed approaches outperform existing state-of-the-art methods and can be used in CPU-based real-time applications. Additionally, we provide a novel dataset comprising images of gymnosperm and angiosperm species.
Dataset and source code are available at
\url{http://github.com/hmarichal93/apd}.

\keywords{Computer vision \and Wood pith detection \and  Deep neural network object detection \and Wood quality}
\end{abstract}
%
%
%


\section{Introduction}
\label{sec:intro}

\begin{figure}[ht]
\begin{center}
   \begin{subfigure}{0.23\textwidth}
   \includegraphics[width=\textwidth]{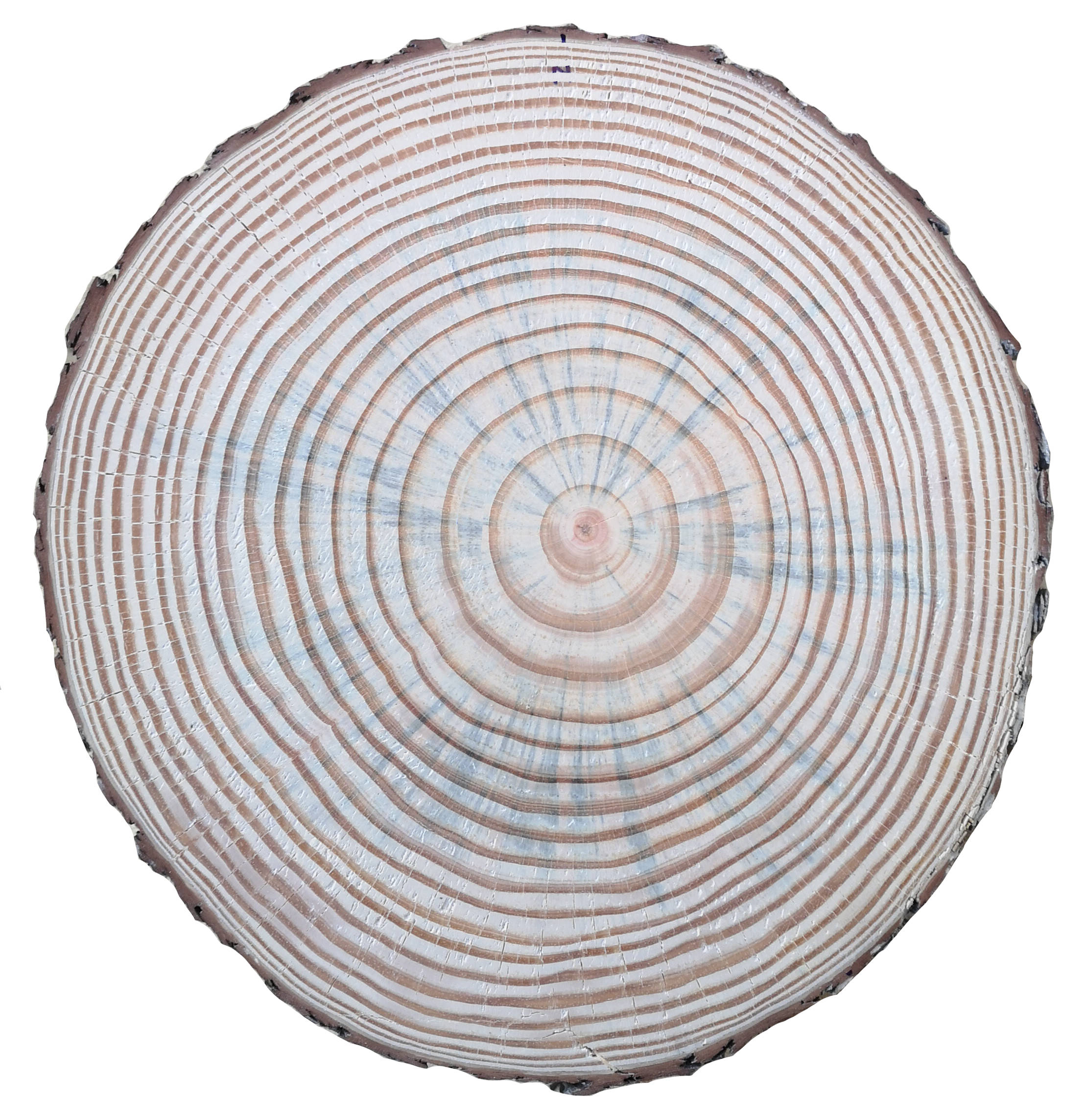}
   \caption{F02d}
   \label{fig:F02d}
   \end{subfigure}
   \begin{subfigure}{0.23\textwidth}
   \includegraphics[width=\textwidth]{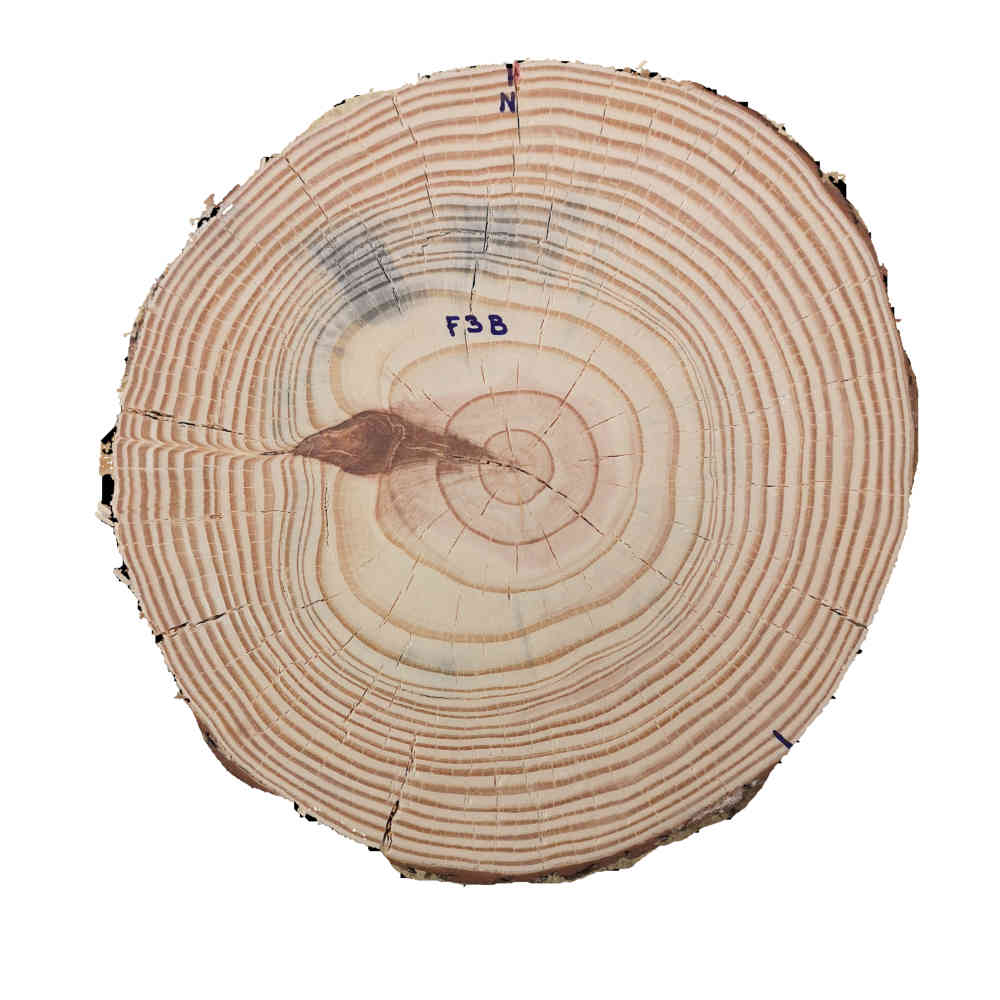}
   \caption{F03d}
   \label{fig:f03d_segmented}
   \end{subfigure}
   \begin{subfigure}{0.23\textwidth}
   \includegraphics[width=\textwidth]{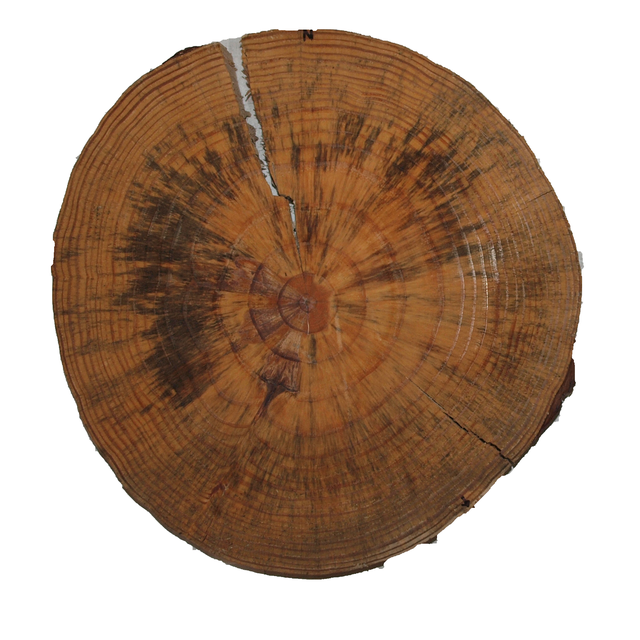}
   \caption{F07e}
   \label{fig:F07e}
   \end{subfigure}
    \begin{subfigure}{0.23\textwidth}
   \includegraphics[width=\textwidth]{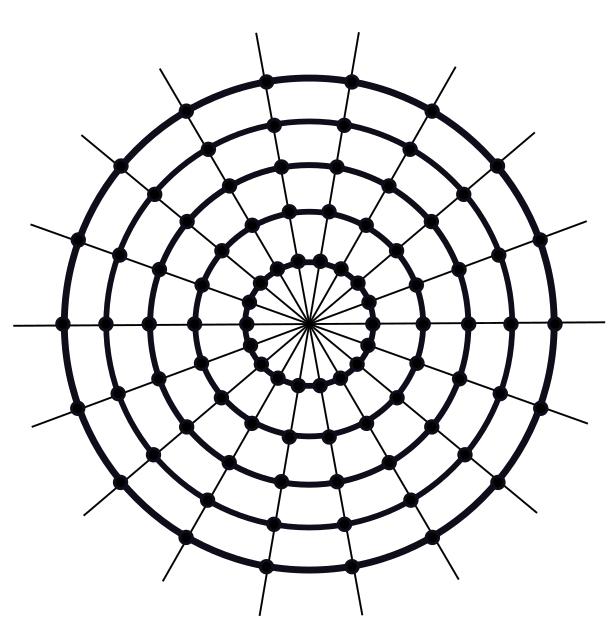}
   \caption{Spider web}
   \label{fig:spiderweb}
   \end{subfigure}
   \hfill
   \caption{(a) to (c) Some examples from \textbf{UruDendro} 
   dataset\cite{UruDendro}, (d) The whole structure, called \textit{spider web}, is formed by a \textit{center} (the slice pith), \textit{rays}, and the \textit{rings} (concentric curves). In the scheme, the \textit{rings} are circles, but in practice, they can be (strongly) deformed as long as they don't intersect another \textit{ring}.} 
   \label{fig:MainIdea}
\end{center}
\end{figure}

Locating the pith of tree cross-sections is essential to identify (in basal discs) the first year of growth and, therefore, the tree's age. The pith has a different type of tissue than the rest of the tree, with distinct physical-mechanical properties. Locating the pith is useful, among other reasons, to detect growing eccentricity; because in the natural process of senescence of standing trees, the fungi that degrade the wood enter through the pith or because the industry discards that part as it has different uses than the rest of the wood. Moreover, some tree ring delineation algorithms are sensitive to a precise pith location \cite{CerdaHM07,JacobiSets,  marichal2023cstrd, Norell-2009}, mainly when those algorithms are based on the ring structure, a concentric pattern similar to a spider web as illustrated in  \Cref{fig:MainIdea}. That figure shows some examples of the diversity of images of tree slices. Ideally, the intersection point between the perpendicular lines through the tree rings should be the pith (the center of the structure, located inside the medulla of the tree). The \textit{spider web} model is only a general approximation, as it is depicted in \Cref{fig:MainIdea}.d. Real slices include ring asymmetries, cracks, knots, fungus, etc., as seen in \Cref{fig:MainIdea}.b and  \Cref{fig:MainIdea}.c. 
Different species produce diverse patterns. Moreover, gymnosperm, as the ones illustrated in  \Cref{fig:MainIdea}, and angiosperm species produce a different wood structure, as seen in \Cref{fig:bbdd}.e. Automatic pith detection must be robust to such variations and perturbations. 

This paper presents several key contributions: the release of a new challenging dataset (UruDendro2 and UruDendro3) for wood pith detection, the development of real-time automatic detection methods (APD and APD-PCL), training of a YoloV8 net (APD-DL) for the same purpose, and rigorous comparison with state-of-the-art methods on various public datasets. These contributions enhance the field of wood pith detection, offering practical solutions and insights for real-time applications.

\section{Previous work}
\label{sec:antecedents}
Schraml and Uhl et al. \cite{Schraml2013} proposed a method (here called LFSA) that splits the wood cross-section into patches, estimating the patch's orientation by 2D Fourier Transform. They accumulate the patch's orientation using a Hough Transform approach and estimate the pith position as the maximum in the accumulation space. 

Kurdthongmee et al. \cite{Kurdthongmee2018} proposed the Histogram Orientation Gradient to estimate the tree ring local orientation and proceed similarly to \cite{Schraml2013}. 
In the same line, Norell et al. \cite{NORELLpith} proposed two ways for estimating the local orientations: quadrature filters and a Laplacian pyramids approach. 
Recently, Decelle et al. \cite{Decelle2022} proposed ACO, a method based on an ant colony optimization algorithm for the local orientation accumulation step. 

Deep Neural Network (DNN) methods have also been applied to solve this problem. Kurdhongmeed et al. \cite{KURDTHONGMEE2020} compared the effectiveness of two DNN object detector models (YoloV3 and SSD MobileNet) to locate the pith. They trained the models via transfer learning over 345  wood slice RGB images captured within a sawmill environment and evaluated over a separate dataset of 215 images.

\section{APD: Automatic Wood Pith Detection}
\label{sec:methodClasico}

We propose an automatic pith detection method based on a model of the wood slice. In a gymnosperm tree cross-section, as the ones shown in \Cref{fig:MainIdea}, two types of structures are present: the rings formed by (roughly) concentric curves and (in some cases) the presence of radial structures such as cracks and fungi. Both are fundamentally related to the pith. The former is due to the growing process of the tree, which forms the rings, and the latter is because the tree's anatomy leads naturally to the radial characteristic of cracks and fungus growing. From this observation derives the principal idea of the proposed method: giving an image of a tree cross-section, we can locate the pith at the intersection of the lines supported by radial structures and the perpendiculars to the rings.

The angiosperm tree cross-section structure is slightly different, as seen in \Cref{fig:ExGleditsia}. Still, it is also formed of radially organized cells, with texture patterns that appear at different radii of the pith. This produces visual macrostructures that allow a similar approach to determine the pith position as depicted in the previous paragraph.

Not always do those hypotheses stand out completely. Sometimes, the ring structure can be highly (locally) deformed, as in the presence of a knot. Sometimes, there are no cracks or fungi present. But in general, enough information is produced by the ring structure and, eventually, by the presence of cracks and fungi to estimate the pith location correctly.

\begin{figure}[ht]
\begin{center}
   \begin{subfigure}{0.23\textwidth}
   \includegraphics[width=\textwidth]{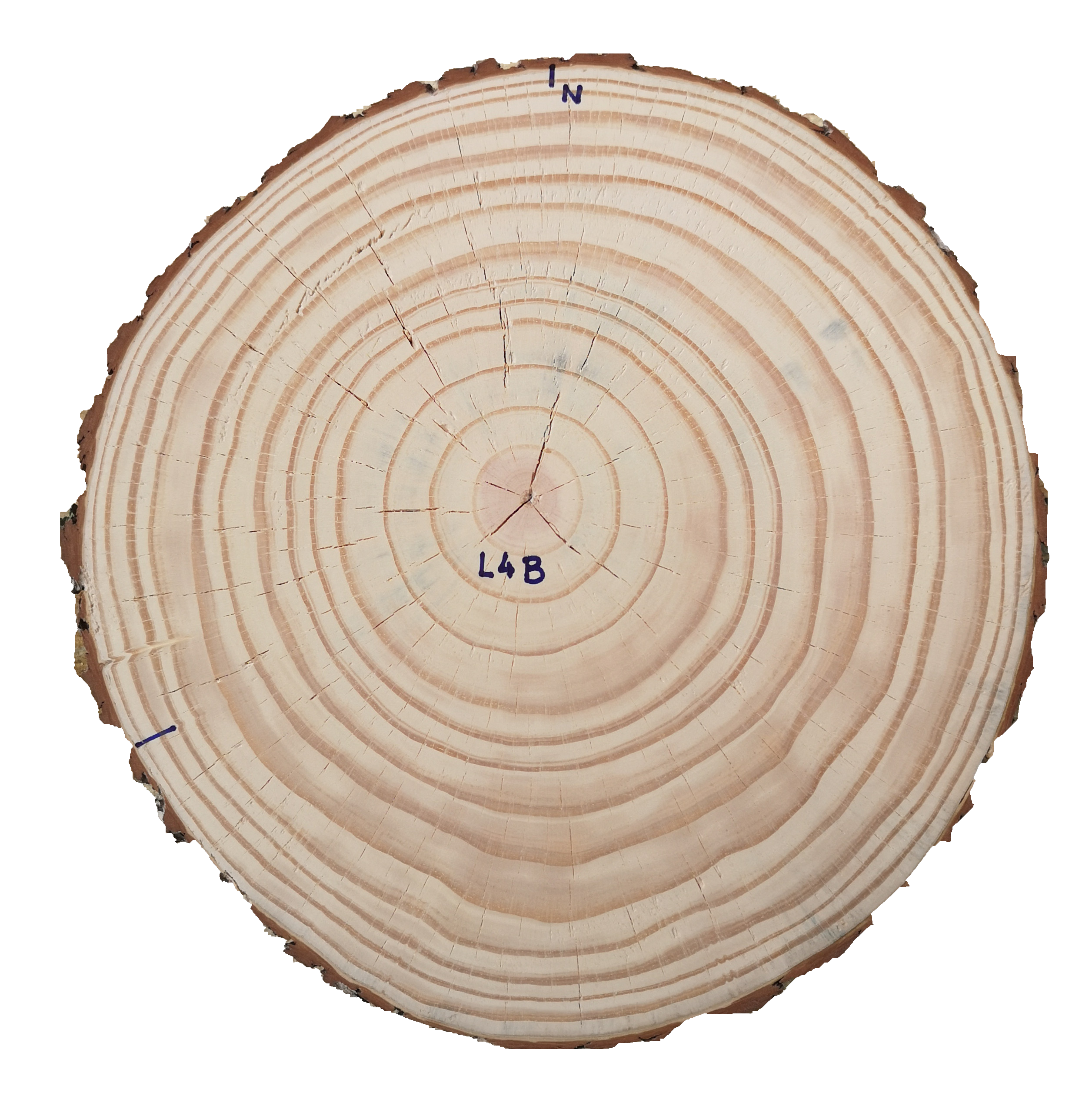}
   \caption{}
   \label{fig:l04d}
   \end{subfigure}
   \begin{subfigure}{0.23\textwidth}
   \includegraphics[width=\textwidth]{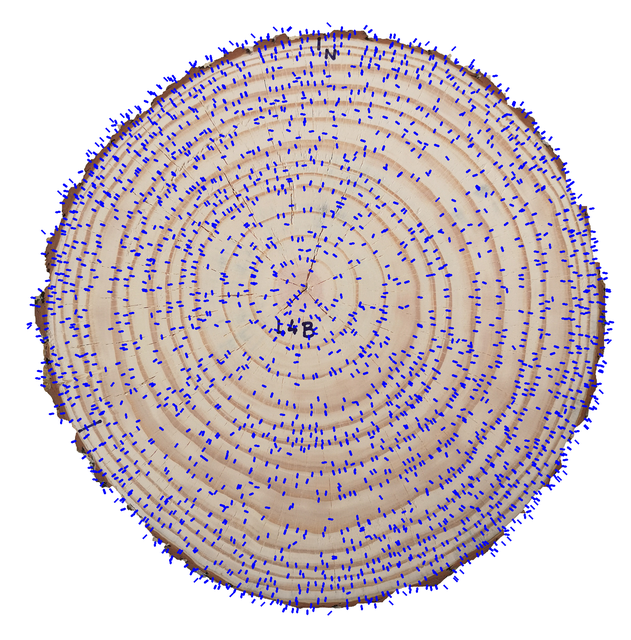}
   \caption{}
   \label{fig:LO}
   \end{subfigure}
   \begin{subfigure}{0.23\textwidth}
   \includegraphics[width=\textwidth]{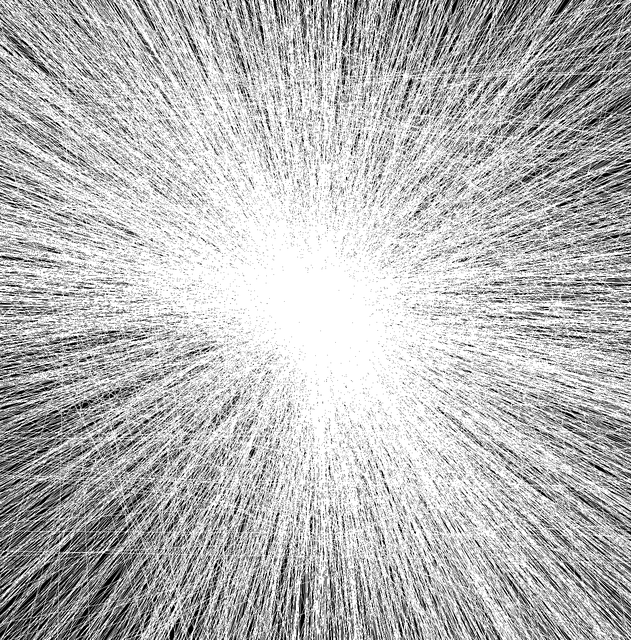}
   \caption{}
   \label{fig:convergingLines}
   \end{subfigure}
   
   \begin{subfigure}{0.23\textwidth}
   \includegraphics[width=\textwidth]{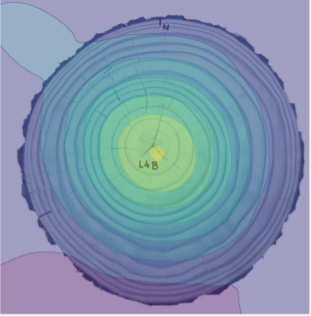}
   \caption{}
   \label{fig:costFunction}
   \end{subfigure}
   \begin{subfigure}{0.23\textwidth}
   \includegraphics[width=\textwidth]{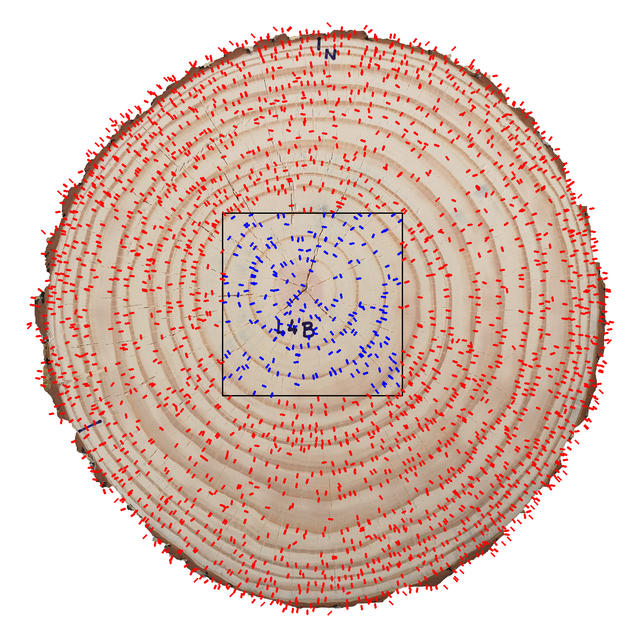}
   \caption{}
   \label{fig:subimage}
   \end{subfigure}
   \begin{subfigure}{0.23\textwidth}
   \includegraphics[width=\textwidth]{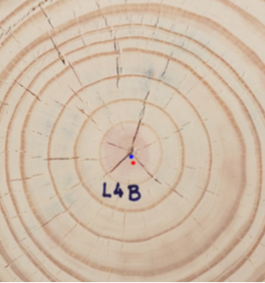}
   \caption{}
   \label{fig:evolution}
   \end{subfigure}
   \hfill
   \caption{Principal steps of APD method (L04d image from UruDendro2 collection). (a) Resized slice image, without background; (b) Sampled LO produced by the Structure Tensor estimation; (c) Accumulation space defined by the LO supported lines; 
   (d) Plot of the cost function (\Cref{eq:cos2}), highest values in yellow; (e) Sub image built around the solution $c_1$ obtained after the first iteration; (f) Evolution of $c_i$. The final solution is in blue; previous iterations' solutions are in red.} 
   \label{fig:method}
\end{center}
\end{figure}

Given an image of the tree cross-section, and using the \textit{spider web} model illustrated in \Cref{fig:MainIdea}.d, the APD approach pseudocode is described at \Cref{algo:APD}. The main steps are the following (see ~\Cref{fig:method} for more details):

\begin{algorithm}
    \caption{APD}
        \label{algo:APD}
    \KwIn{$Im_{in}$, //\text{RGB slice image}\;
    }
    \KwOut{Pith location}
    $ST_{O}$, $ST_{C}$ $\leftarrow$ local\_orientation($Im_{in}$, $st_\sigma$, $st_{w}$) \\
    $LO_f$ $\leftarrow$ lo\_sampling( $ST_{O}$, $ST_{C}$, $lo_{w}$, $percent_{LO}$)\\
    $LO_r$ $\leftarrow$ $LO_f$\\
    \For{$i$ \textbf{in} $1$ \textbf{to} $max\_iter$}{      
     \If{ $i$ $>$ 1 }{
        $LO_{r}$ $\leftarrow$ filter\_lo\_around\_$c_i$($LO_f$, $r_{f}$, $c_i$), // See \Cref{fig:method}.e\\
     }
     $c_{i+1}$ $\leftarrow$ optimization($LO_{r}$)     // ~\Cref{eq:optimizationProblem}\\
     \If{$\|c_{i+1} - c_{i}\|_{2}$ $<$ $\epsilon$}{
     $c_i$ $\leftarrow$ $c_{i+1}$\\
     break 
     }
     $c_i$ $\leftarrow$ $c_{i+1}$
    } 
    \KwRet{$c_i$}
\end{algorithm}

\begin{figure}[ht]
\begin{center}
   \includegraphics[width=0.3\textwidth]{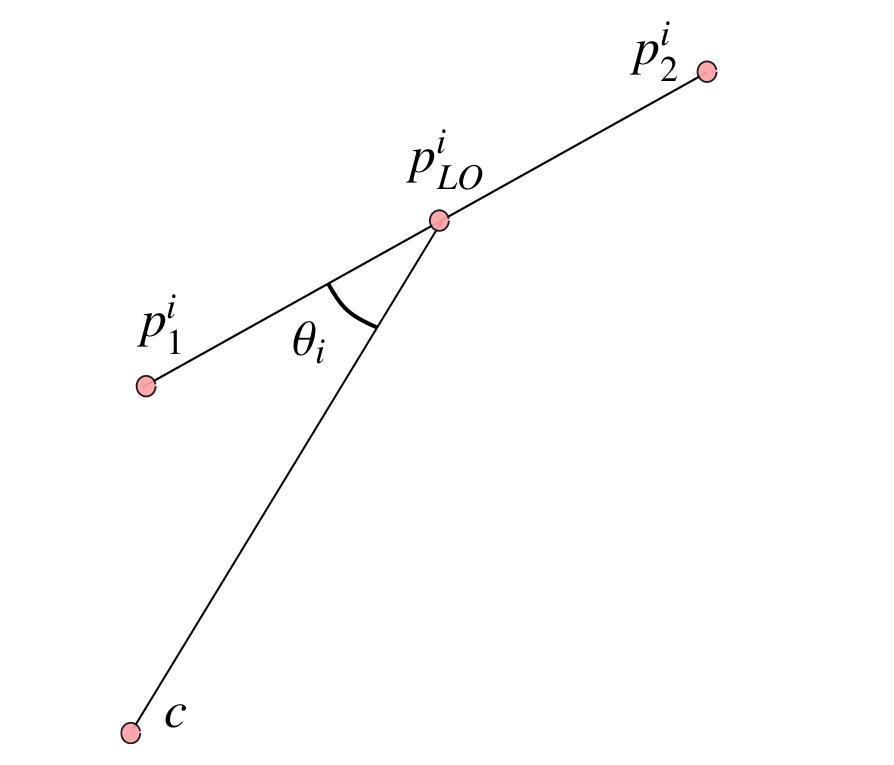}
   \caption{Cost function definitions} 
   \label{fig:CostFunctionDefinitions}
\end{center}
\end{figure}

\begin{enumerate}
  
    \item \textit{Local orientation detection (line 1 of \Cref{algo:APD})}. To estimate the local orientation (LO), we compute the 2D-Structure Tensor \cite{BigunTensorStructure} $ST[p]$ at each pixel $p$ using a window of size \textit{$st_{w} \times st_w$}. Pixels in the window are weighted by a Gaussian kernel $w$ of parameter $st_\sigma$. The  structure tensor is calculated as $ST[p]=\sum_{r}w[r]ST_{xy}[p-r]$ where $ST_{xy}[p]$ is defined as 
    $$
    ST_{xy}[p] =
    \begin{bmatrix}
       (I_x[p])^2     & I_x[p]I_y[p]\\
       I_x[p]I_y[p] & (I_y[p])^2
     \end{bmatrix}
    $$
    
    where $I_x[p]$ and $I_y[p]$ are the first derivatives of image $I$ in point $p$ along $x$ and $y$, respectively. For simplification, we can re-write the $2 \times 2$ structure tensor matrix at pixel p as:
    $$ST[p]=
    \begin{bmatrix}
    J_{11} & J_{12} \\ J_{12} & J_{22}
    \end{bmatrix}
    $$

    The local orientation at pixel p is:

    \begin{equation}
    ST_{O}[p] = \frac{1}{2} \arctan(\frac{2J_{12}}{J_{22}-J_{11}})
    \label{eq:sto}
    \end{equation}

    The \textit{coherence} of the LO estimation in $p$ is given by the relative value of eigenvalues $\lambda_1$ and $\lambda_2$ (where $\lambda_1$ is the largest eigenvalue, and $\lambda_2$ is the smallest one):
    
    \begin{equation}
    ST_{C}[p] = \left(\frac{\lambda_1 - \lambda_2}{\lambda_1 + \lambda_2}\right)^2
    \label{eq:stc}
    \end{equation}

    The outputs of this step are two matrices: one of local orientations ($ST_{O}$) and one of coherence ($ST_{C}$).
   
    \item \textit{Local orientation sampling (line 2 of \Cref{algo:APD})}. The LO estimations are sampled in the following way: 1) $ST_{O}$ and $ST_{C}$  are divided in non-overlapping patches of size $ lo_w \times lo_w$. 2) We find the pixel $p^j$ with the highest coherence ($c^{j}_{high}$) within patch $patch_i$. A minimum patch coherence $st_{th}$ is defined. We assign $ST_O[p^j]$ to $patch_i$ in position $p^j$, if $c^j_{high} > st_{th}$. To fix $st_{th}$, we calculate the value of $ST_{C}$ such that a given percentage (parameter $percent_{LO}$) of the LO in the slice has $ST_{C} > st_{th}$. Each LO is a segment $lo_i = \overline{p^{i}_{1} p^{i}_{2}}$, defined by the limits $p^{i}_1$ and $p^{i}_2$. $p^{i}_{LO}$ is the middle point between them ($p^j$). Given the local orientation  $\alpha_i = ST_{O}[p^{i}_{LO}]$, points $p^{i}_1$ and $p^{i}_2$ are computed as $p^{i}_{1,2} = p^{i}_{LO} \pm (\cos(\alpha_i), \sin(\alpha_i))$. 
    
    Suppose $N$ patches have coherently enough LO; the output of the step is a matrix, $LO_f$ of size $N\times 4$. In this way, lines are supported by the LO of all meaningful structures in the cross-section, such as the rings.
    
    \item \textit{Find the center (line 7 of \Cref{algo:APD})}. Given the filtered local orientation matrix, $LO_{f}$, we define the following optimization problem: find $c_{opt}$, the geometrical position that maximizes the collinearity between the $lo_i$  and a line passing by $c_{opt}$ and $p^i_{LO}$. To this aim, we define the following cost function: 
    
    \begin{equation}
    h(x,y) = \frac{1}{N}\sum_{i=1}^{N}{\cos^{2}(\theta_{i}(x,y)})
    \label{eq:cos2} 
    \end{equation}
    
    \Cref{fig:CostFunctionDefinitions} illustrates the vectors involved in computing \Cref{eq:cos2}.
    The angle between  $\overline{p^{i}_{1} p^{i}_{2}}$ and $\overline{c p^{i}_{LO}}$ is $\theta_i$. The pith position $c$ of coordinates ($x$,$y$) is the origin of a segment $\overline{c p^{i}_{c}}$. 
     As $\cos(\theta_{i}(x,y)) = \frac{<\overline{c p^{i}_{LO}}, \overline{p^{i}_{1} p^{i}_{2}}>}{|\overline{c p^{i}_{LO}}||\overline{p^{i}_{1} p^{i}_{2}}|}$, the optimization problem to be solved becomes: 
    
    \begin{equation}
    \begin{aligned}
    c_{opt} = \max_{c} \quad & 
    \frac{1}{N} \sum_{i=1}^{N}
    \left(\frac{<\overline{c p^{i}_{LO}}, \overline{p^{i}_{1} p^{i}_{2}}>}{|\overline{c p^{i}_{LO}}||\overline{p^{i}_{1} p^{i}_{2}}|}\right)^2 \\
    \textrm{s.t.} \quad & c \in Slice\hspace{0.1cm}Region\\
    \end{aligned}
    \label{eq:optimizationProblem}
    \end{equation}

    \item To find the global maximum ($c_{opt}$) of the former optimization problem, we use the \textit{minimize} method from \textit{scipy.optimize} python package with its default parameters. 
    Problem \ref{eq:optimizationProblem} is convex if it is restricted to the region of the wood cross-section. Therefore, as initialization to the \textit{minimize} method, we use the least squares solution of finding the point $c_{ini}$, which minimizes the distance to all the lines in $LO_r$ within the slice.
        
    \item \textit{Refinement (lines 4 to 11 of \Cref{algo:APD}}). Once a candidate for the pith location $c_{opt}$ is obtained, the optimization procedure (\ref{eq:optimizationProblem}) is repeated using only the local orientations within a squared region of size $Size_{image}/r_{f}$ centered in $c_{opt}$ (see \Cref{fig:method}.e). This step is repeated until the pith location doesn't move more than a given tolerance ($\epsilon = 10^{-5}$) or the iteration counter reaches $max\_iter = 5$. This approach avoids distortions introduced by a (very) asymmetric tree ring growth pattern.  
\end{enumerate}

\section{APD-PCL: PClines based Automatic Wood Pith Detection}
\label{sec:methodClasicoRobusto}

The APD method described in \Cref{sec:methodClasico} works fine when the ring structure gives enough information. In some (rare) cases, the ring structure is not visible due to fungi or other perturbations. In those cases, it is possible to solve the same problem using the lines supported by the radial structure of those perturbations in addition to the lines produced by the ring structure. For this reason, the APD-PCL version of the method is more robust and allows for the successful treatment of cross-sections with highly degraded ring patterns. The price to pay is a slower algorithm, as it includes a RANSAC-based clustering step.

The APD-PCL method selects which local orientations to consider in the optimization problem of \Cref{eq:optimizationProblem}. 
In general, the estimation made by the structure tensor calculation step is determined by the rings. Different perturbations also produce some LO, but its number is minimal, and the lines they support don't converge to the pith. In some (rare) cases, the perturbations are so important that they overshadow the ring structure. In those cases, the number of LO produced by the perturbations is more significant than those produced by the ring structure. The set of perturbations-related local orientations can be of diverse origin: knots, fungi, cracks, and noise. Some of them (namely fungi and cracks) have a typical radial orientation, so the perpendicular lines to its LO converge to the pith. 

\begin{figure}[ht]
\begin{center}
   \begin{subfigure}{0.23\textwidth}
   \includegraphics[width=\textwidth]{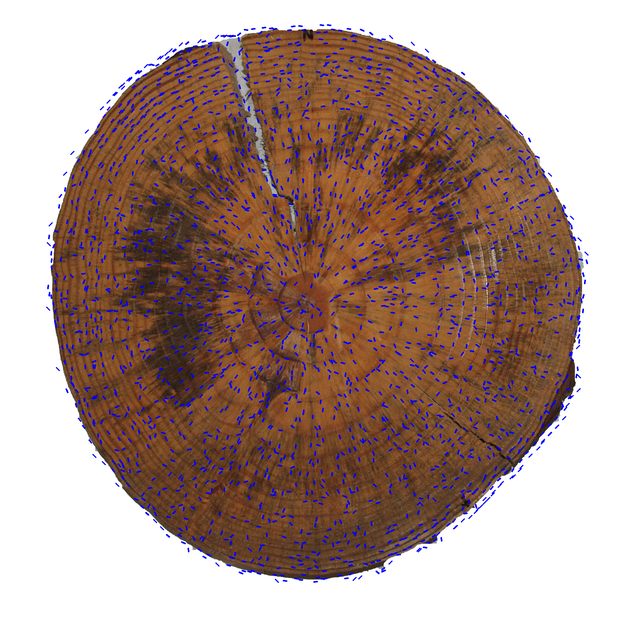}
    \caption{}
   \label{fig:TS-PClines}
   \end{subfigure}
   \begin{subfigure}{0.23\textwidth}
   \includegraphics[width=\textwidth]{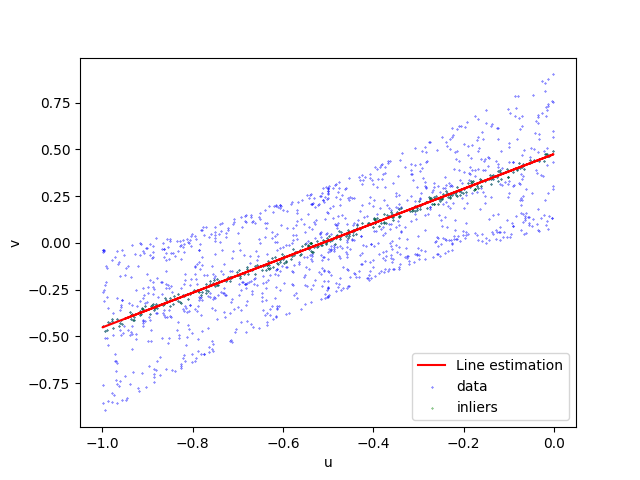}
   \caption{}
   \label{fig:RotatedPClines}
   \end{subfigure}
   \begin{subfigure}{0.23\textwidth}
   \includegraphics[width=\textwidth]{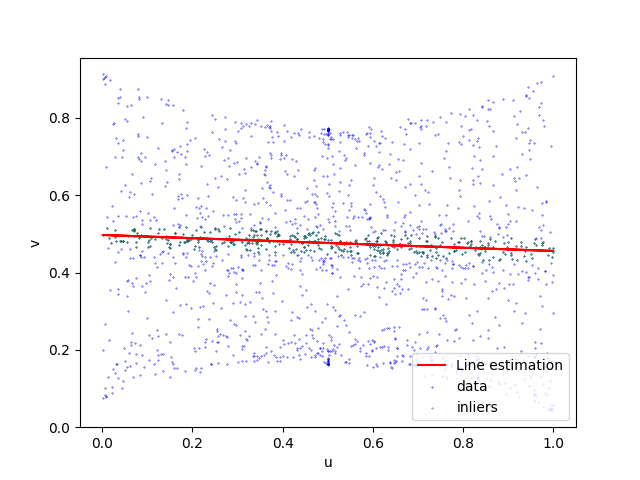}
   \caption{}
   \label{fig:clusterStraight}
   \end{subfigure}
   \begin{subfigure}{0.23\textwidth}
    \includegraphics[width=\textwidth]{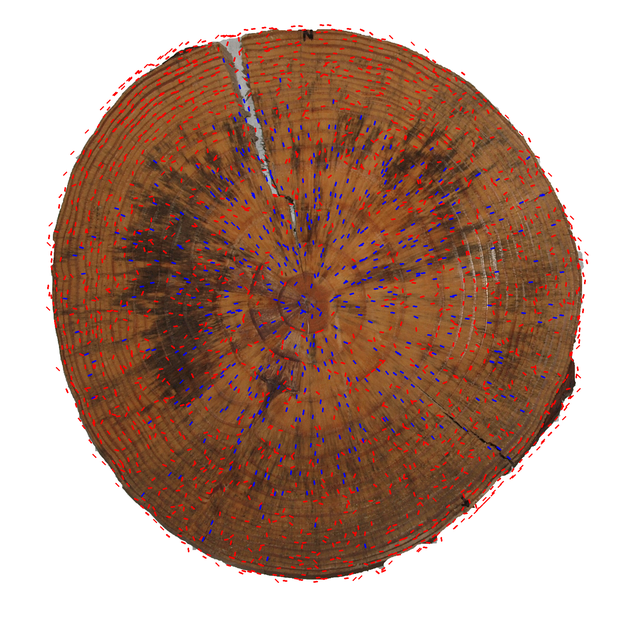}
   \caption{}
   \label{fig:ConvergingLinesPClines}
   \end{subfigure}
   \hfill
   \caption{Use of PClines to cluster converging local orientations 
   for slice F07e (same as \Cref{fig:F07e}). (a) Local orientations; (b) Selection of the converging segments in the twisted space using RANSAC  to fit a line (in red). Inliers are colored in green; (c) The same procedure is applied in the straight space; (d) In blue, the converging LO (inliers from both sub-spaces) and the LO to be removed in red.} 
   \label{fig:PClines}
\end{center}
\end{figure}

Considering this, we modify \Cref{algo:APD}, by including a post-processing step over matrix $LO_f$, between lines 2 and 3. The rest of the algorithm is the same: 

\begin{enumerate}
    \item Use the PClines transform \cite{PClines} to convert each line into a point. 

    \item The PClines space is formed by two sub-spaces defined by a parameter $d$: the \textit{straight space} includes lines with orientations $\alpha_i \in [0,\frac{\pi}{2}$] and the \textit{twisted space} lines with orientations $\alpha_i \in [\frac{\pi}{2},\pi]$. As seen in \Cref{fig:PClines}, convergent lines in the Euclidean space correspond to aligned points in the PClines spaces. This allows the following steps:
    
    \begin{enumerate}
        \item Lines supported by LO produced by the ring structure converge somewhere around the pith. They produce a line-shaped cluster in the PClines spaces (figures \ref{fig:RotatedPClines} and \ref{fig:clusterStraight}). We select the aligned points using a RANSAC \cite{ransac} approach. We work only in the $[-d,0]$ and $[0,d]$ ranges for the twisted and straight sub-spaces, respectively. This avoids the use of points near the infinity. We select all the converging lines in both spaces, excluding those simultaneously selected in both sub-spaces.
        
        \item The previous step clusters all convergent $LO_f$ in the image producing the set $LO_{ring}$. We rotate by $90$ degrees all the orientations in $LO_f$ and repeat the previous procedure to detect the converging ones. These rotated converging lines cluster, $LO_{radial}$, is produced by cracks, fungi, or similar structures. Adding both gives the set:
        
        $$LO_f^{PClines} = LO_{ring} + LO_{radial}$$
        \label{item:step_rotated}.
        
        \item To make the line segment selection method more robust, we add a third PClines transform using the lines supported by $LO_f^{PClines}$. Most ring-related LO and rotated LO generated by radial structures are expected to converge (hence, to form a line cluster in the PCline space). Therefore, most of the outliers should be removed at this step.
    \end{enumerate}
\end{enumerate}    

\begin{figure}[ht]
\begin{center}
   \begin{subfigure}{0.23\textwidth}
   \includegraphics[width=\textwidth]{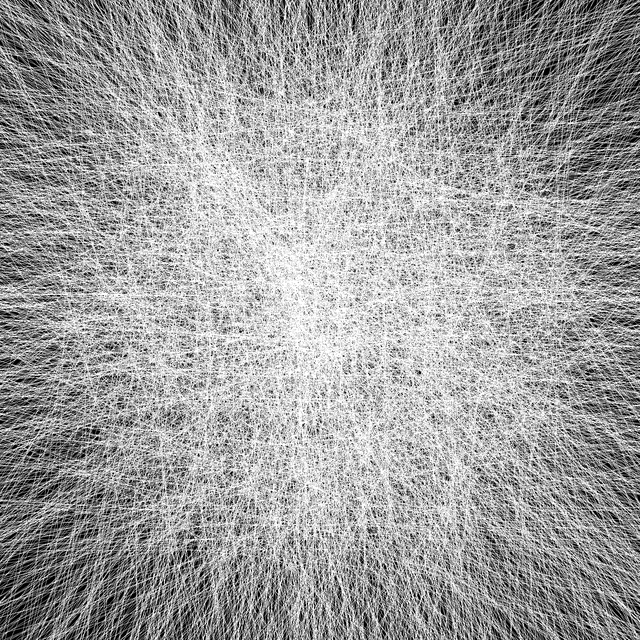}
   \caption{}
   \label{fig:F07e_lo}
   \end{subfigure}
   \begin{subfigure}{0.23\textwidth}
   \includegraphics[width=\textwidth]{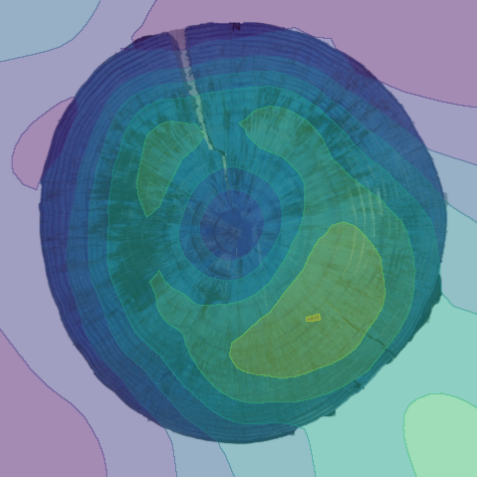}
   \caption{}
   \label{fig:F07e_cost}
   \end{subfigure}
   \begin{subfigure}{0.23\textwidth}
   \includegraphics[width=\textwidth]{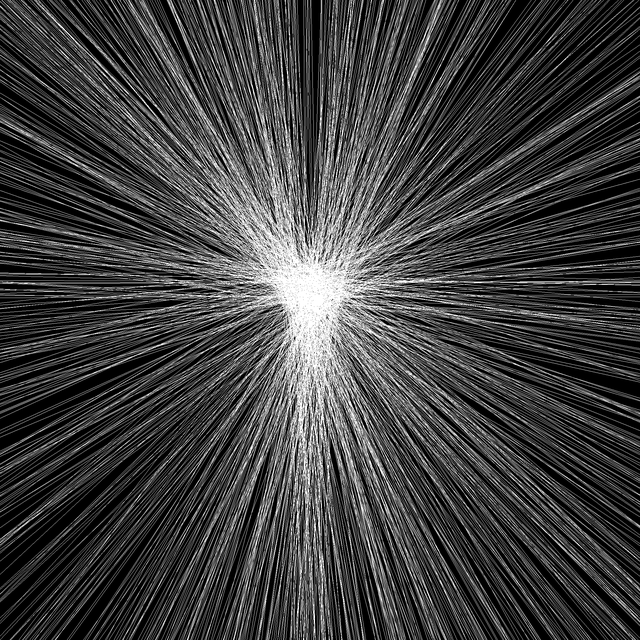}
   \caption{}
   \label{fig:F07e_PClines_lo}
   \end{subfigure}
   \begin{subfigure}{0.23\textwidth}
   \includegraphics[width=\textwidth]{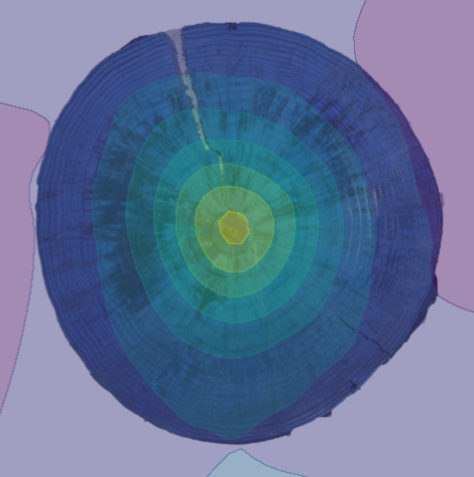}
   \caption{}
   \label{fig:F07e_PClines_cost}
   \end{subfigure}
   \hfill
   \caption{LO Accumulation space and cost function for slice F07e with and without applying the PClines filtering method. (a) LO Accumulation space with no filtering; (b) cost function of LO with no filtering; (c) LO Accumulation Space with PClines filtering; (d) cost function of LO with PClines filtering} 
   \label{fig:PClines_example}
\end{center}
\end{figure}

~\Cref{fig:PClines_example} illustrates the considered lines and the accumulation space of \Cref{eq:cos2} without and with the PClines step. Note how the method filters out many non-convergent lines and regularizes the cost function.  

The APD-PCL method is similar to the APD one, but the PClines-based filtering step diminishes the number of considered lines, filtering out many non-convergent ones.

\section{APD-DL: Deep Learning based Automatic Wood Pith Detection}
\label{sec:methodML}

In Sections \ref{sec:methodClasico} and \ref{sec:methodClasicoRobusto}, we tackle the pith detection problem using a (\textit{spider web}) model, as in the "classic" image processing times. Now, we present a Deep-learning approach that learns the model from the data. To this aim, Kurdthongmee et al. \cite{Kurdthongmee2019} used a YoloV3 model. Inspired by them, we train a YoloV8 \cite{YOLOV8} network using the datasets described in \Cref{sec:bbdd}. This is an architecture tailored for object detection and segmentation. To train it, we must provide a set of data formed by wood cross-section images labeled with the ground truth position of the pith as a bounding box.

We use a five-fold cross-validation technique. We divide the data into five sets. In each fold $i$, we use one set ($test_i$) for testing and the other four for training. The training process in each fold is done as usual, and we use the produced model to label the data in $test_i$. The process is repeated for all the folds. In the end, we have predictions for all the data, and in each case, the used model was generated without the influence of the $test_i$ data.
With the predictions for all images produced in this manner, we can deliver the metrics to determine the method's performance. 

\begin{table}[ht]
\centering
\begin{tabular}{lccccc}
\hline
Collection  & Mean (Std)  & Median  & Max & FN    \\ \hline
Uru2        & 0.55 (1.45)  & 0.18   & 11.32 & 2	 \\ 
Uru3        & 0.13 (0.06)   & 0.13  & 0.27  & 0    \\ 
Kennel      & 0.14 (0.07)  & 0.13   & 0.24 & 0     \\ 
Forest      & 0.45 (1.85)  & 0.12   & 13.91 & 0	 \\  
Logyard     & 0.52 (1.29)   & 0.27   & 7.51 & 0	   \\ 
Logs        & 0.22 (0.46)   & 0.13   & 4.42 & 1	 \\ 
Discs       & 0.23 (0.54)   & 0.14   & 5.67	& 0  \\ 
All         & 0.33 (1.01)   & 0.14   & 13.91 & 3  \\ \hline
\end{tabular}
\caption{Prediction results of 5-fold cross-validation for the APD-DL. The second to fourth columns show the Mean (and standard deviation in parenthesis), Median, and Maximum normalized error (defined in \Cref{sec:normalizedErrors}) values. The last column shows the false negatives. We use all the datasets together for train the model and calculate the performance within each dataset.}
\label{tab:resultsAPD-ML}
\end{table}

\Cref{tab:resultsAPD-ML} show the results using normalized errors (see \Cref{sec:normalizedErrors}). Training with such a high diversity of data produces state-of-the-art results. 
Results over each row (collection) are calculated using the predictions produced during the five-fold cross-validation with all the images in the seven datasets. In some (rare) cases, this approach doesn't give a prediction (hence a false negative). In those situations, the method gives the center of the image as the pith position. This explains the (relatively) large value of the Maximum error and the differences between the Mean and Median errors. Besides the rare false negatives, the results are excellent. The last row depicts the results for all the collections. 

\paragraph{Hyperparameters} The algorithm was trained with images of width size $640$ (keeping the aspect ratio), using a batch size of $16$, for $100$ epochs, with the optimizer AdamW \cite{loshchilov2019decoupled} ($lr=0.002$, $momentum=0.9$) and yolov8n as pre-trained weights. All the network was re-trained.

\section{Datasets description}
\label{sec:bbdd}

\begin{figure}[ht]
\begin{center}
   \begin{subfigure}{0.23\textwidth}
   \includegraphics[width=\textwidth]{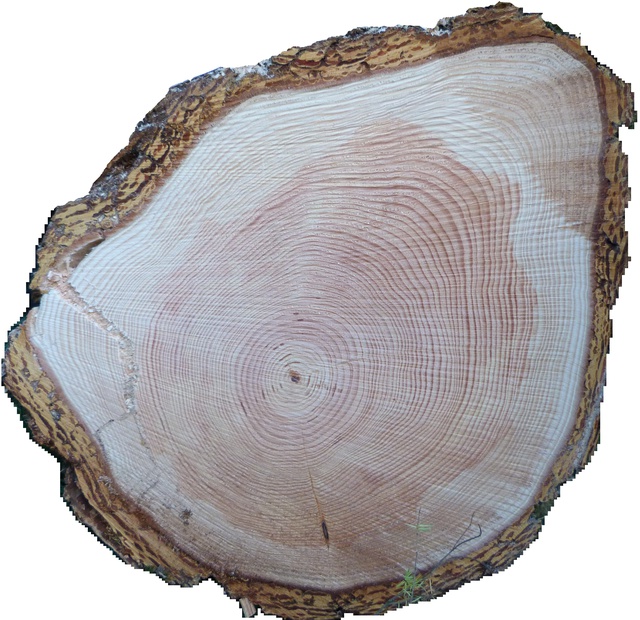}
   \caption{Forest}
   \label{fig:ExForest}
   \end{subfigure}
   \begin{subfigure}{0.23\textwidth}
   \includegraphics[width=\textwidth]{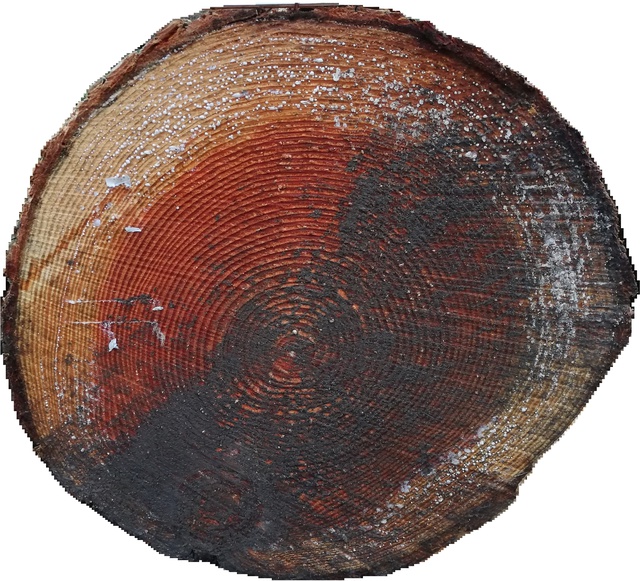}
   \caption{Longyard}
   \label{fig:ExLongyard}
   \end{subfigure}
   \begin{subfigure}{0.23\textwidth}
   \includegraphics[width=\textwidth]{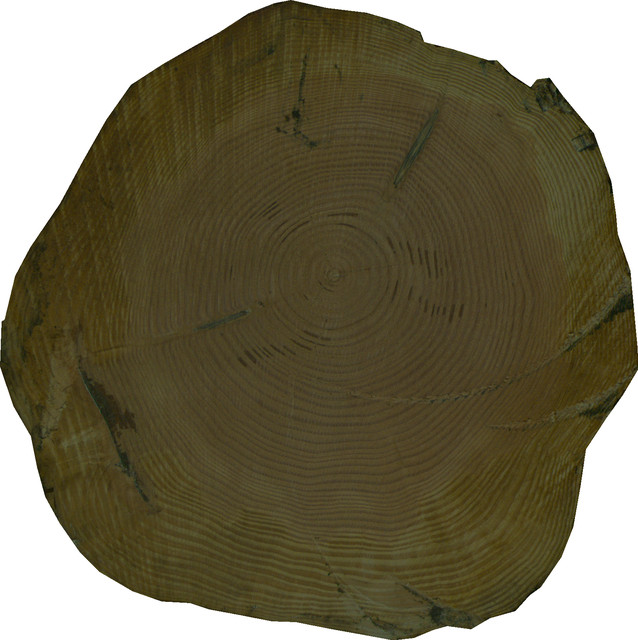}
   \caption{Logs}
   \label{fig:ExLogs}
   \end{subfigure}
   
   \begin{subfigure}{0.23\textwidth}
   \includegraphics[width=\textwidth]{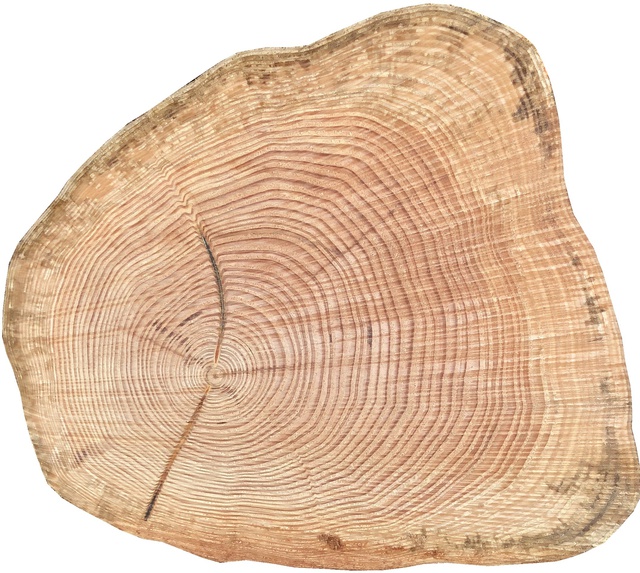}
   \caption{Disc}
   \label{fig:ExDisc}
   \end{subfigure}
   \begin{subfigure}{0.23\textwidth}
   \includegraphics[width=\textwidth]{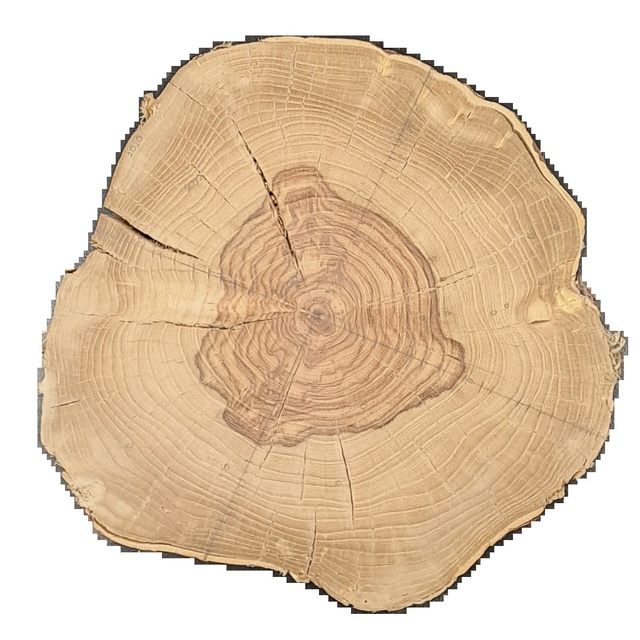}
   \caption{Uru3}
   \label{fig:ExGleditsia}
   \end{subfigure}
   \begin{subfigure}{0.23\textwidth}
   \includegraphics[width=\textwidth]{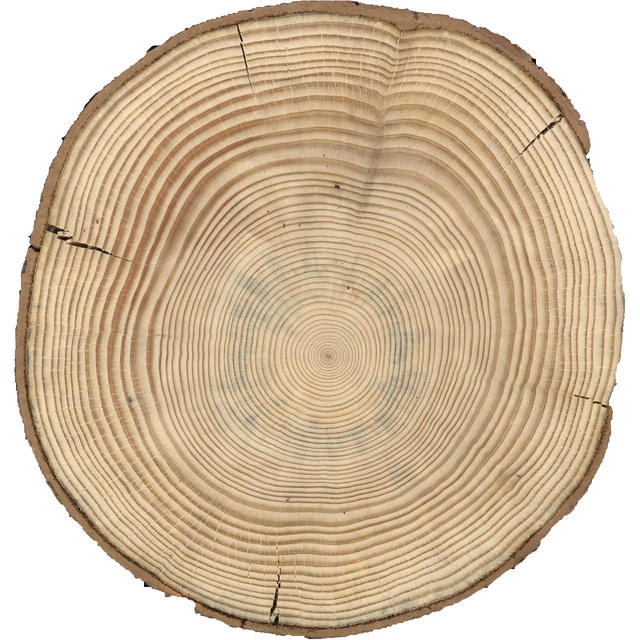}
   \caption{Kennel}
   \label{fig:ExKennel}
   \end{subfigure}
   \hfill
   \caption{Examples of the used datasets. Species are (a-d) \textit{Douglas fir}, (e) \textit{Gleditsia triacanthos}, (f) \textit{Abies alba}. Acquisition conditions: (a-c) in the field, with a smartphone camera; (d-e) in the laboratory, with controlled illumination. The samples of images (d-e) were previously sanded and polished. The samples of images (a-c) didn't have any special treatment.} 
   \label{fig:bbdd}
\end{center}
\end{figure}

We use the following datasets:
\begin{itemize}
    \item UruDendro. 
    We introduce here a new public dataset with two collections of wood cross-section samples with experts annotated ground truth \cite{UruDendro}:
    \begin{itemize}
        \item UruDendro2: 119 RGB images of \textit{Pinus taeda}  slices. This collection includes 64 images taken under different illumination conditions and cameras, published in 2022 in our website \cite{UruDendro},
        increased with 55 new images. 
        The new images were taken in laboratory conditions, with an iPhone 6S phone (12 Mpx camera) at a distance between 43 and 51 cm from the slice, under controlled illumination with a led ring of 35 W. Size images range between 1000 and 3000 pixels in width.
        The surface of the slices presented different conditions: some were cut by chainsaw, smoothed by a handheld planner, and polished with a rotary sander. 
        All these images are annotated by at least one expert with the position of the pith. 
        \item UruDendro3: 9 RGB images of \textit{Gleditsia triacanthos}, an angiosperm, acquired in a laboratory, without illumination-controlled conditions, using a Huawei P20 Pro smartphone (24 Mpx camera) at a distance of approximately 1 meter from the slice. Size images range between 1000 and 2000 pixels in width. 
        All the slices were polished.
        All these images are annotated by at least one expert with the position of the pith. 
    \end{itemize}    

    \item Kennel \cite{KennelBS15}. A public dataset with 7 RGB 1280 pixels squared images of \textit{Abies alba}, polished and acquired in controlled illumination laboratory conditions. The pith pixel location is provided as metadata.

    \item TreeTrace \cite{longuetaud:hal-03658479}. A public dataset, with samples of \textit{Douglas fir} taken at different stages of the wood process chain. The pith pixel location is provided as metadata. Each image has several wood slices. To build the collections, we extract sub-images containing one slice each, producing almost squared images between 1000 and 3000 pixels in width. This dataset includes the following collections:
    \begin{itemize}
        \item Forest, 57 RGB images taken from the freshly cut logs with a digital camera. 
        \item Logyard, 32 RGB images of the same log ends,  acquired with a smartphone in the sawmill courtyard several days after the cutting. 
        \item Logs, 150 RGB images acquired in the sawmill with a smartphone.
        \item Discs, 208 RGB images acquired with a 400 dpi scanner from sanded and polished slices after several weeks of air-drying. 
   \end{itemize} 

\end{itemize}

\begin{table}[ht]
\centering
\begin{tabular}{lcc}
\hline
Collection & Size & Specie    \\ \hline
 UruDendro2 & 119 & \textit{Pinus taeda} \\  
 UruDendro3 & 9 & \textit{Gleditsia triacanthos} \\ 
 Kennel & 7 & \textit{Abies alba} \\ 
 Forest & 57 & \textit{Douglas fir}  \\  
  Logyard & 32 & \textit{Douglas fir} \\ 
  Logs & 150 & \textit{Douglas fir} \\ 
  Discs & 208 & \textit{Douglas fir} \\ \hline
\end{tabular}
\caption{Dataset description.}
\label{tab:bbdd}
\end{table}

\Cref{tab:bbdd} summarize the used datasets. \Cref{fig:MainIdea} show images from the UruDendro2 dataset, and \Cref{fig:bbdd} show examples from the other collections. These datasets convey a high degree of variability. It includes examples of gymnosperm (\textit{Pinus taeda}, \textit{Abies alba} and  \textit{Douglas fir}) as well as angiosperm (\textit{Gleditsia triacanthos}). Acquisition conditions are also diverse, including images obtained with a smartphone in the forest or the sawmill. Samples were acquired in the field with dirt, sap, or saw marks, and others were obtained in controlled illumination conditions in the laboratory from polished samples. The samples include perturbations as the presence of fungi, cracks, and knots, as can be seen in \Cref{fig:MainIdea} and sap and saw marks, as can be seen in \Cref{fig:bbdd}.b and \Cref{fig:bbdd}.d. 
All have the ground truth position of the pith. Considering all datasets, we work with 582 images.

\section{Results and discussion}
\label{sec:results}
\begin{table*}[ht]
\centering
\begin{tabular}{lccccccc}
\hline
 &  UruDendro2  & UruDendro3 & Kennel & Forest & Logyard & Logs & Discs\\ \hline
LFSA \cite{Schraml2013} & 1.03 (0.85)  & 1.46 (0.97)& 0.42 (0.18)   & 0.80 (0.36)  & 1.02 (0.62)  & 0.80 (0.46)  & 0.72 (0.43) \\ 
ACO \cite{Decelle2022}& 2.23 (6.64)  & 4.52 (11.96)  & 0.2 (0.06)  & 0.24 (0.24)  & 0.60 (1.11)  & 0.46 (0.45) & 0.24 (0.35)\\ 
APD-PCL &  \textbf{0.42 (0.34)} & 0.74 (0.54) & 0.19 (0.10) & 0.81 (0.98) & 0.82 (0.84) & 0.52 (0.47) & 0.46 (0.57) \\ 
APD &  1.02 (2.45) & 0.55 (0.30) & \textbf{0.14 (0.06)} & \textbf{0.22 (0.18)} & \textbf{0.35 (0.17)} & 0.29 (0.33) & 0.26 (0.42)   \\ 
APD-DL &  0.55 (1.45)  & \textbf{0.13 (0.06)}  & 0.14 (0.07)  &  0.45 (1.85)  & 0.52 (1.29)  & \textbf{0.22 (0.46)}  & \textbf{0.23 (0.54)}  \\ \hline
\end{tabular}
\caption{Results on all the datasets. Normalized errors. We show the mean error and the standard deviation between parenthesis.}
\label{tab:resultsPerformance}
\end{table*}
\subsection{Preprocessing}
\label{sec:preprocessing}
Sometimes, the images are acquired in the field with a smartphone camera, and one image can contain more than one cross-section. Regardless of the method used, all images are preprocessed in such a way as to standardize the image input:

\begin{enumerate}
  \item \textit{Background substraction}. Produce a new image limited to one slice. To this aim, when possible, we filter out the background using the mask provided in the datasets. If the mask is not provided, we use a  deep learning-based method \cite{salientObject}, which uses an $U^2Net$ to segment salient objects.  

    \item \textit{Resize the image}. This step, not strictly necessary, allows us to fix the algorithm's parameters once and for all. All images are resized to $640$ pixels width, respecting the original image's aspect ratio. 
\end{enumerate}

\subsection{Normalized errors} 
\label{sec:normalizedErrors}

Given the cross-sections' diverse dimensions, presenting the errors in pixels is not informative. Additionally, not all the datasets provides the millimeter pixel relation. We use the percentage of the equivalent slice radius. Given a prediction $P_i$ and a Ground Truth $GT_i$, this error is calculated as follows:
$$
Err_i = \frac{100 \times Dist(P_i,GT_i)}{Equivalent\_radio(image_i)} 
$$

Where $Dist(P_i,GT_i)$ is the Euclidean distance, in pixels, between the prediction and the Ground Truth. Remember that within this work, the pith is modeled as a point in the image. Therefore $Dist(P_i,GT_i)$ is a distance between points.  $Equivalent\_radio(image_i)$ (in pixels) is half the biggest horizontal or vertical image size that circumscribes the slice, without background, generated by the preprocessing.

\subsection{Experiments}
\label{sec:experiments}

The method to fine-tune the APD-DL method is explained in \Cref{sec:methodML}. To determine the best parameters' values for LFSA, APD and APD-PCL we minimize overall used datasets the average of Euclidean distances between ground truth and predictions. 

For the APD and  APD-PCL methods, the parameters $st_{\sigma}$, 
$r_{f}$ and 
$ransac\_outlier\_th$ were set after experiments over a few images. 
Therefore, they are considered as fixed. The rest of the parameters, $percent_{LO}$,  $st_{w}$ and $lo_{w}$ were set following the procedure described in the former paragraph. The search of the minimum was over the following grid:  $percent_{LO}$ in $[0.3, 0.5, 0.7, 0.9]$,  $st_{w}$ in $[3, 7, 9, 11]$ and $lo_{w}$ in $[3, 7, 9, 11]$.

Inferences were made using an Intel Core i5 10300H workstation with 16GB and a GPU GTX1650 (when needed).

\subsection{Results} 
\label{sec:results_sb}
 
In this section, a performance comparison is made between the different methods. \Cref{tab:resultsPerformance} show the performance of the proposed methods and two state-of-the-art ones \cite{Decelle2022, Schraml2013}, over the datasets presented in \Cref{sec:bbdd}.  We use the mean error and standard deviation, using normalized errors, to compare different-size wood cross-sections. The performance of the methods differs for each collection due to its specific characteristics regarding species, acquisition conditions, etc. Note that ACO was developed (and tailored) for the TraceTree collections. Its performance degrades when tried on other species (such as UruDendro collections). LFSA performance is more regular across collections. The three methods proposed in this paper outperform ACO and LFSA on all collections. APD and APD-DL perform better for almost all collections. APD outperforms APD-PCL for all collections except UruDendro2, which has some images with fungi and cracks overshadowing the ring structure. Note that in all the cases, the precision of the pith detection is very high.

\begin{table}[ht]
\centering
\begin{tabular}{lcccccc}
\hline
Method &  Mean   &  Median & Max    &  FN & Time   \\ \hline
LFSA \cite{Schraml2013}   & 0.83 	& 0.72  &	5.03 &  0 & 627\\ 
ACO \cite{Decelle2022} & 0.79 	&0.21   &	36.39 & 2 & 918 \\ 
APD-PCL &  0.52 	& 0.34  &	4.33 & 0 & 2339 \\ 
APD & 0.42 & 0.19  & 15.44	 & 0 &  784\\ 
APD-DL & 0.33	& 0.14  &	13.91  & 3 & 209\\ \hline
\end{tabular}
\caption{Results of all the methods over the whole set of images, i.e., merging all collections. Normalized errors, number of false negatives, and execution time in milliseconds.}
\label{tab:resultsAll}
\end{table}

 \Cref{tab:resultsAll} compares the performance of all tested methods using the 582 images of all collections. All methods presented in this paper outperform LSFA and ACO methods. The APD performance is surpassed only by the APD-DL method but at the cost of some false negatives: images in which APD-DL didn't find a solution. We can see that APD slightly outperforms the APD-PCL method. This is due to the RANSAC algorithm used to cluster points in the PClines space. When there is no clear clustering of points around a line, RANSAC tries to fit a line anyway, selecting a wrong set of LO and producing a wrong pith localization. This situation sometimes appears in the TreeTrace dataset.
 To consider the mean processing time per image for each method, it must be considered that APD-DL and ACO methods run on GPU, while APD, APD-DL, and LFSA run on a CPU machine. Note that APD is roughly three times faster than APD-PCL. All in all, it is remarkable that the "classic style" model-based proposed methods (APD and APD-PCL) and a Deep Learning one (APD-DL) have similar performance and execution times, allowing real-time applications with a CPU in the APD and APD-PCL cases.
In the supplementary material, we add showcases illustrating how the different methods work under extreme conditions. 
\section{Conclusions and future work}
\label{sec:conclusions}
This paper addresses the wood pith detection on tree slices problem using classic image processing and machine learning-based approaches. Both approaches are determined by the characteristics of the data used to tune the algorithm. In search of a more general solution, we use a set of diverse datasets, which spans different species, acquisition conditions, and perturbations (from cracks and knots to saw marks and dirt for images acquired on the field).

We proposed three real-time methods. The first two are based on a \textit{spider web} model in a classic image processing approach, and the third one is a Deep Learning method. The former has excellent performance, runs in real-time on a CPU-based machine, and the model allows a clear comprehension of the approach. The limited number of parameters is understandable and can be fixed once and for all. The latter has better (although similar) performance but has some false negatives and is more opaque concerning the meaning of its millions of parameters. Moreover, it runs on a GPU based machine.

The UruDendro dataset, with annotated images of \textit{Pinus taeda} (a gymnosperm) and \textit{Gleditsia triacanthos} (an angiosperm), are presented and can be used by the community to test other approaches to this problem. 


\section{Supplementary material}

\subsection{APD Parameters}
\label{sec:parameters}

 The number of parameters in the machine learning approaches is huge, and they don't have a clear physical meaning. In the classic approaches, such as APD and APD-PCL methods, a limited number of parameters are included, and they have a physical or algorithmic meaning.  
 The APD method has the following parameters, fixed once and for all after a grid search with all the datasets. The default values are in parentheses:

\begin{itemize}
    \item $st_{\sigma}$: Structure tensor Gaussian $\sigma$ ($1.2$). 
    \item $st_w$: Structure tensor Gaussian kernel size ($3$ for APD and $7$ for APD-PCL). 
    \item $percent_{LO}$: to fix $st_{th}$, the minimum coherence value to consider valid an LO ($0.7$).
    \item $lo_{w}$: LO sampling window size ($3$ for APD and $7$ for APD-PCL).
    \item $r_{f}$: a factor used to estimate the side size of the search region for the iteration ($7$). 
\end{itemize}

The APD-PCL method adds the following:

\begin{itemize}
\item $ransac\_outlier\_th$: RANSAC residual threshold defining the width of the line cluster ($0.03$).
\end{itemize}

\subsection{Methods comparison: difficult cases}

\begin{figure}[h]
\begin{centering}
    \begin{subfigure}{0.25\textwidth}
    \begin{centering}
    \includegraphics[width=\textwidth]{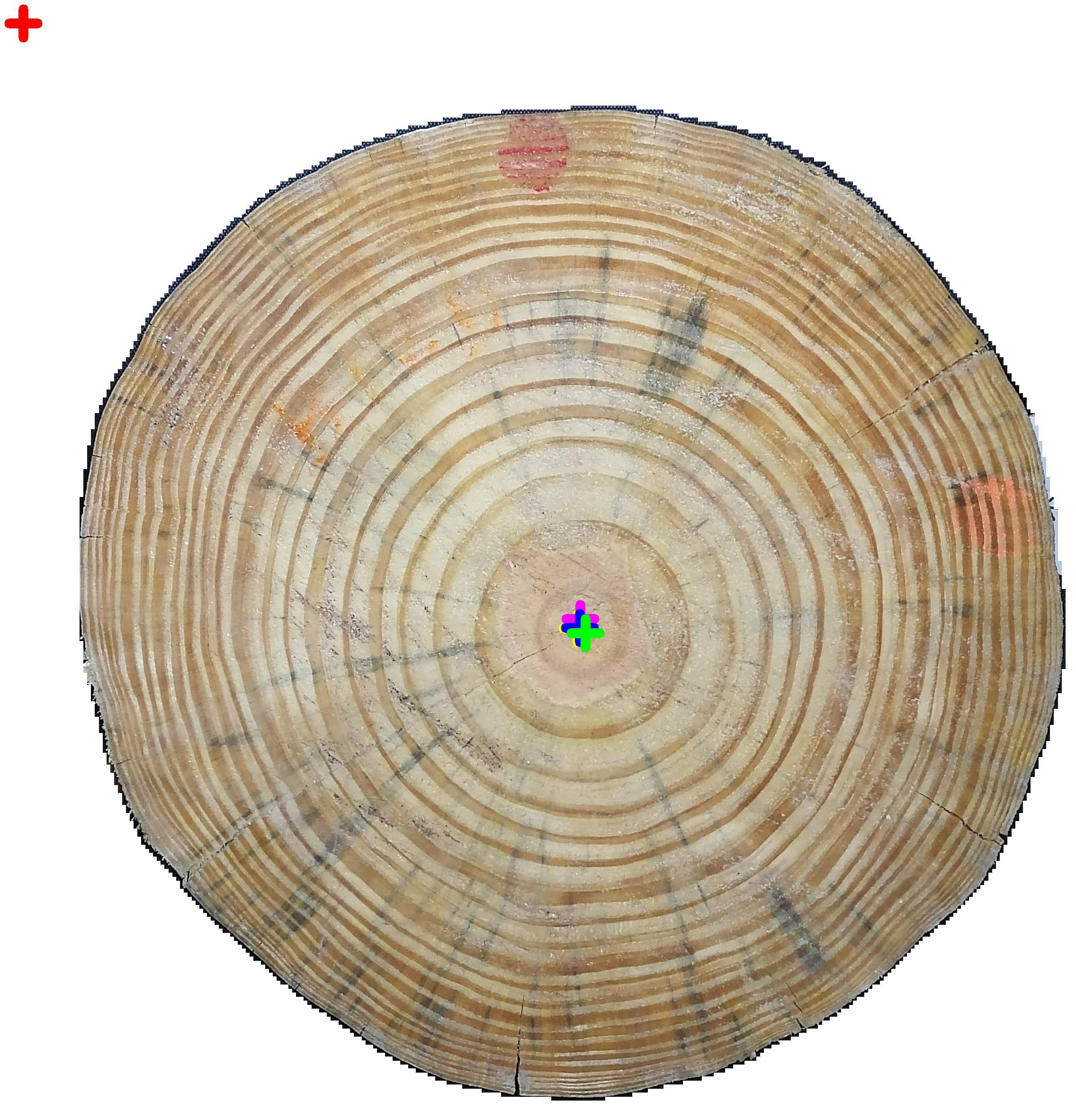}
    \label{fig:B12-2}
    \caption{B12-2}
    \end{centering}
    \end{subfigure}
    \hfill
    \begin{subfigure}{0.25\textwidth}
    \begin{centering}
    \includegraphics[width=\textwidth]{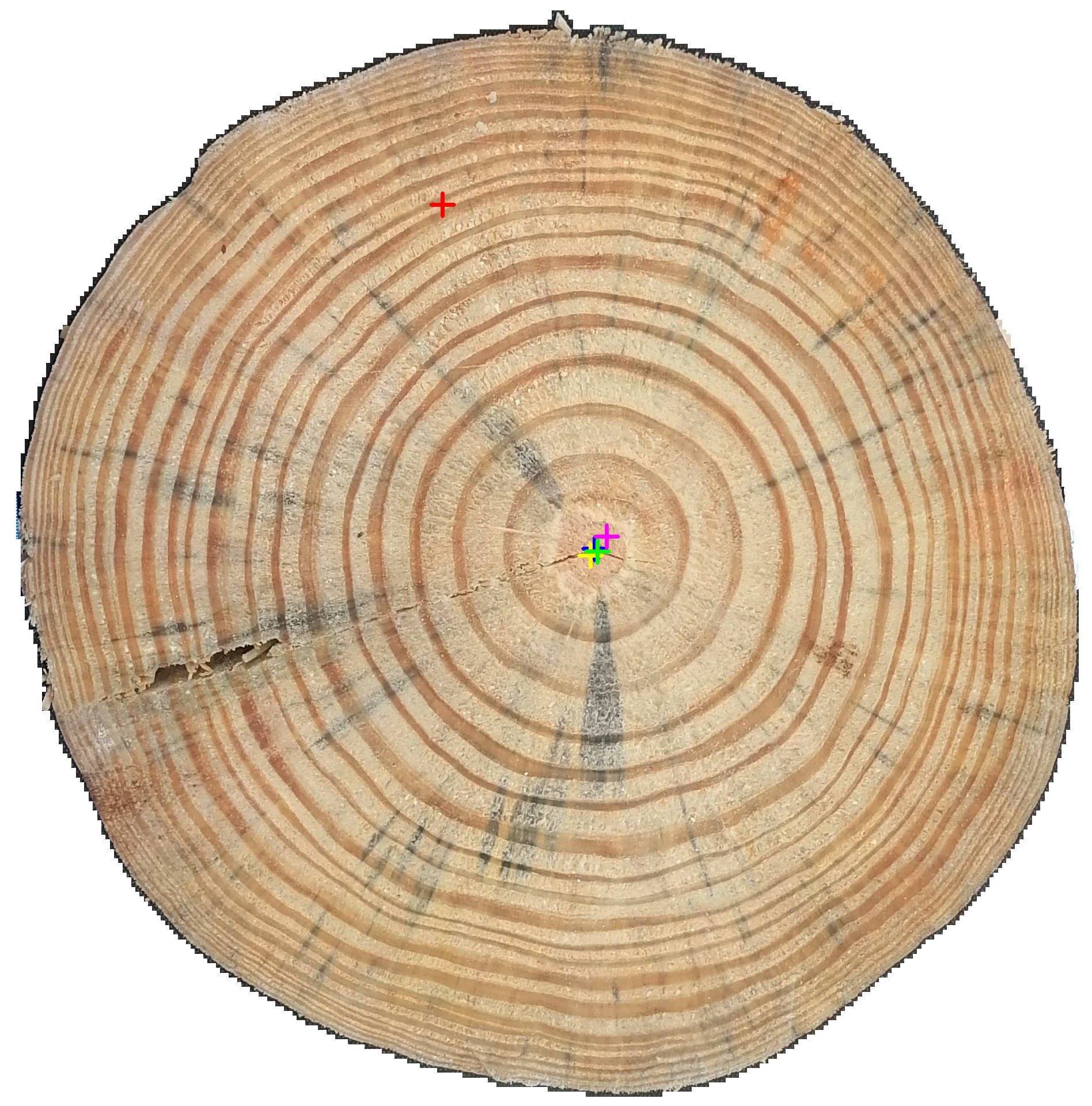}
    \label{fig:C17}
    \caption{C17}
    \end{centering}
    \end{subfigure}
    \hfill
    \begin{subfigure}{0.25\textwidth}
    \begin{centering}
    \includegraphics[width=\textwidth]{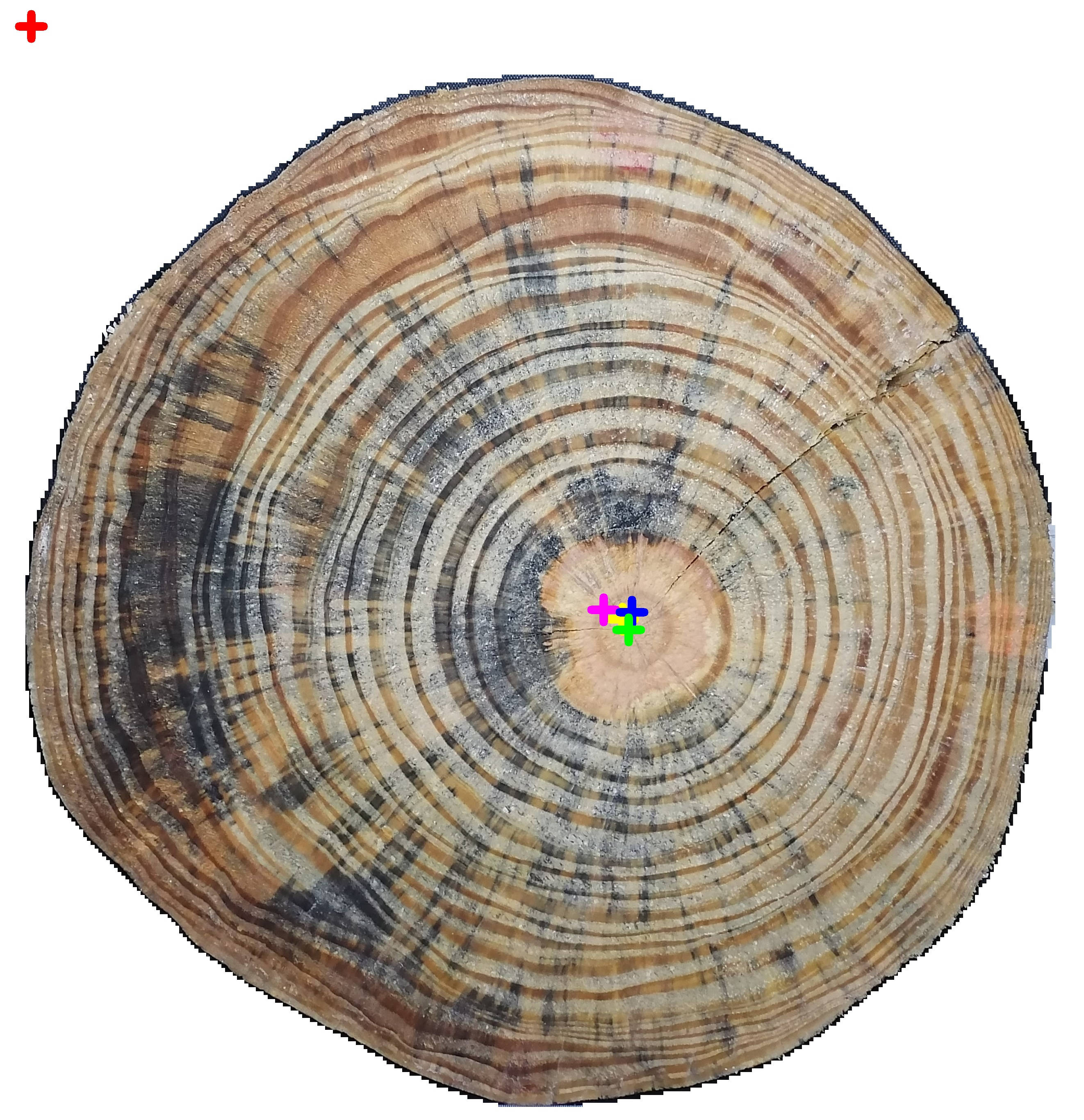}
    \label{fig:C10-2}
    \caption{C10-2}
    \end{centering}
    \end{subfigure}
    \hfill
    
    \begin{subfigure}{0.25\textwidth}
    \begin{centering}
   \includegraphics[width=\textwidth]{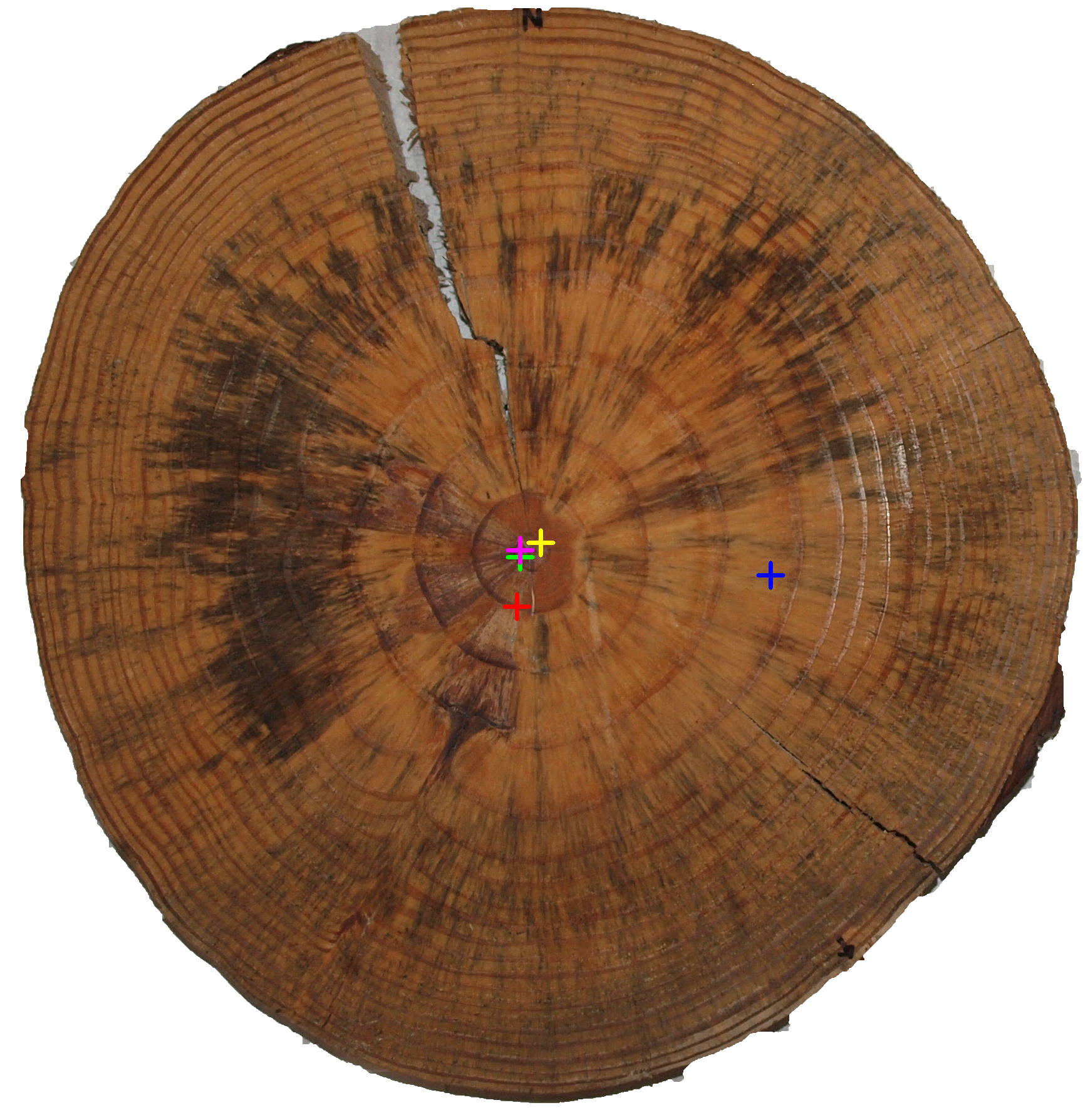}
    \label{fig:F07e_d}
    \caption{F07e}
    \end{centering}
    \end{subfigure}
    \hfill
    \begin{subfigure}{0.25\textwidth}
    \begin{centering}
   \includegraphics[width=\textwidth]{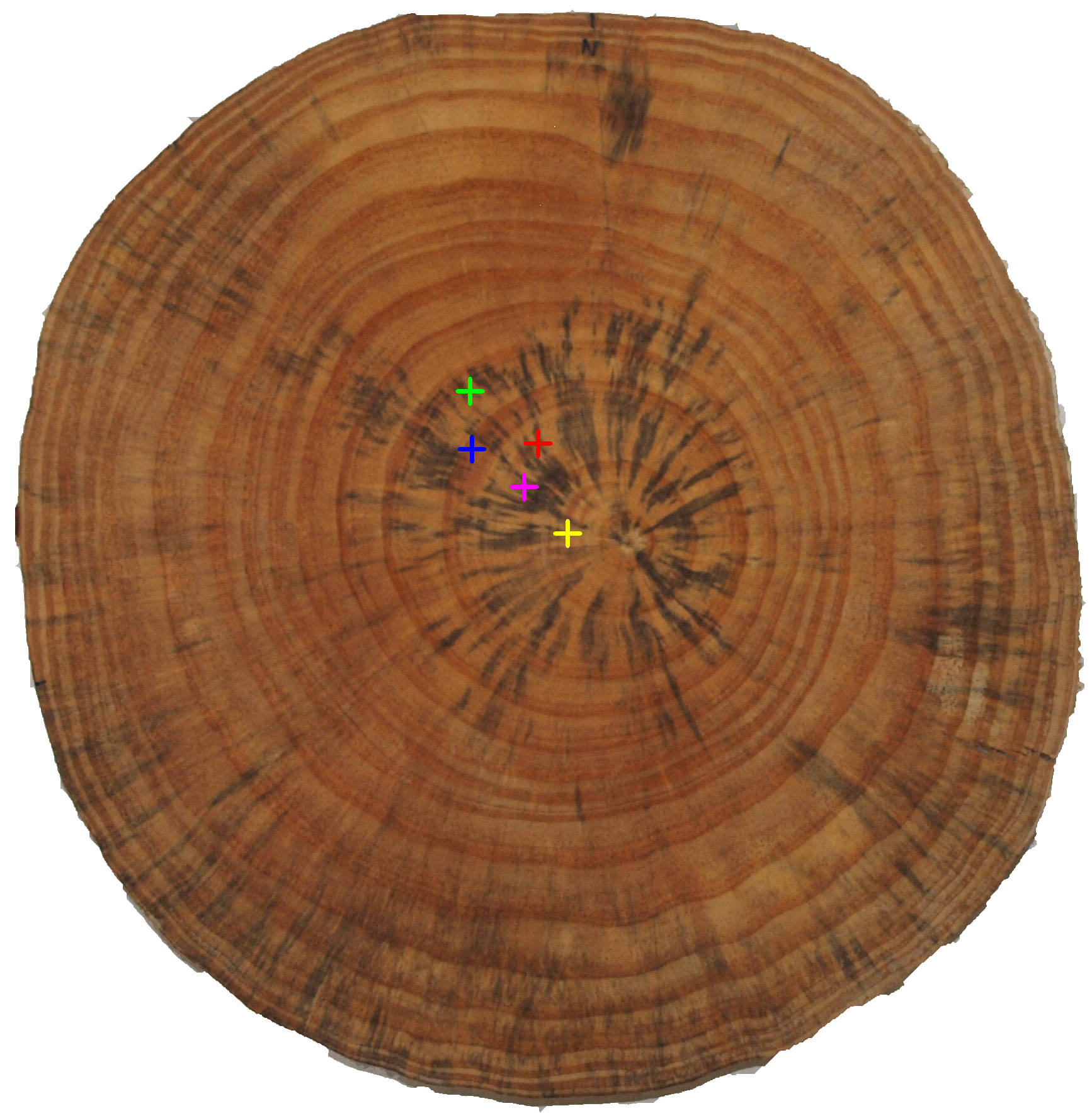}
    \label{fig:L02a}
    \caption{L02a}
    \end{centering}
    \end{subfigure}
    \hfill
    \begin{subfigure}{0.25\textwidth}
    \begin{centering}
   \includegraphics[width=\textwidth]{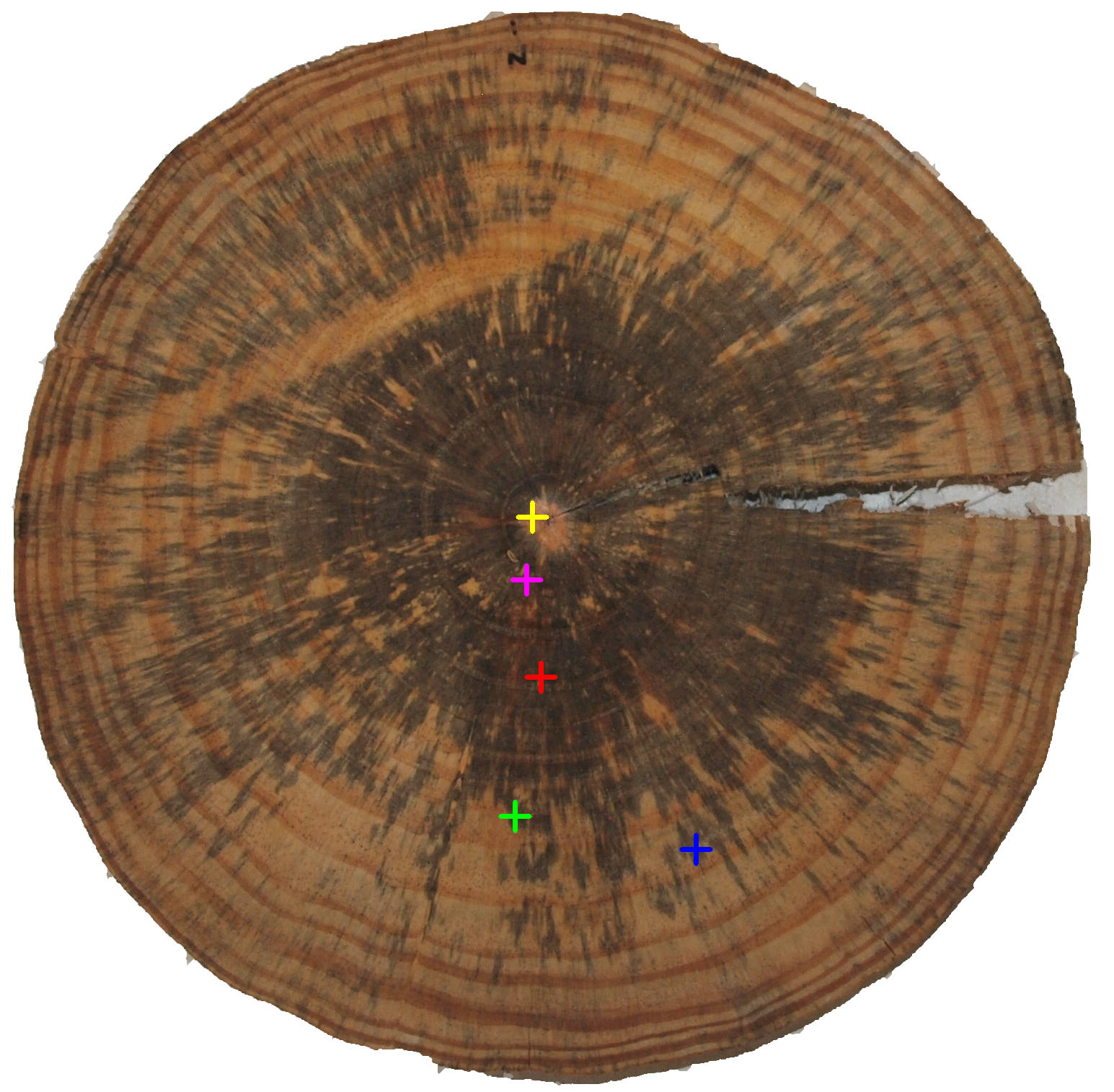}
    \label{fig:L02b}
    \caption{L02b}
    \end{centering}
    \end{subfigure}
    \hfill

   \caption{Results over difficult cases, e.g cracks, fungus presence in Uru2 collection. Purple, LFSA; Red, ACO;  Blue, APD; Yellow, APD-PCL and Green, APD-DL }
   \label{fig:discussion_worst_cases}
\end{centering}
\end{figure}

\begin{figure}[ht!]
\begin{centering}
    \begin{subfigure}{0.25\textwidth}
    \begin{centering}
    \includegraphics[width=\textwidth]{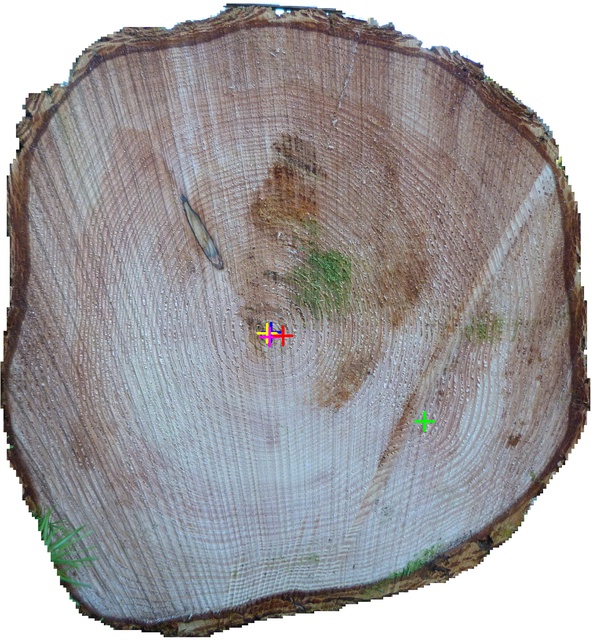}
    \label{fig:D02-L1-9}
    \caption{D02-L1-9}
    \end{centering}
    \end{subfigure}
    \hfill
    \begin{subfigure}{0.25\textwidth}
    \begin{centering}
    \includegraphics[width=\textwidth]{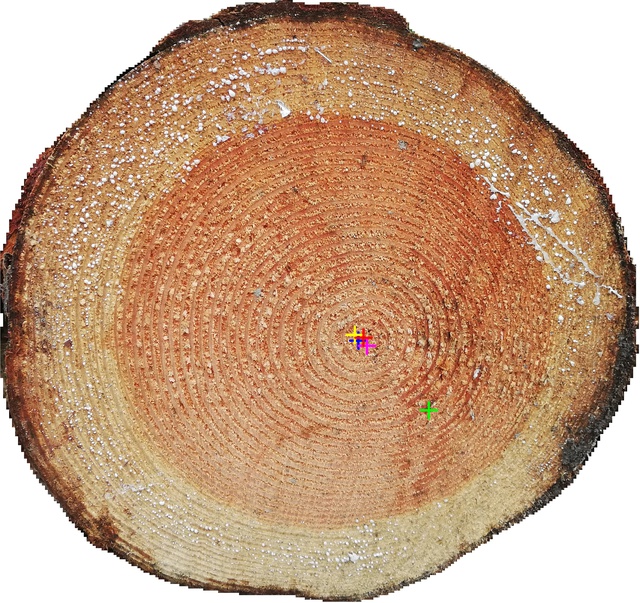}
    \label{fig:D01-L4-BBF-2}
    \caption{D01-L4-BBF-2}
    \end{centering}
    \end{subfigure}
    \hfill
    \begin{subfigure}{0.25\textwidth}
    \begin{centering}
    \includegraphics[width=\textwidth]{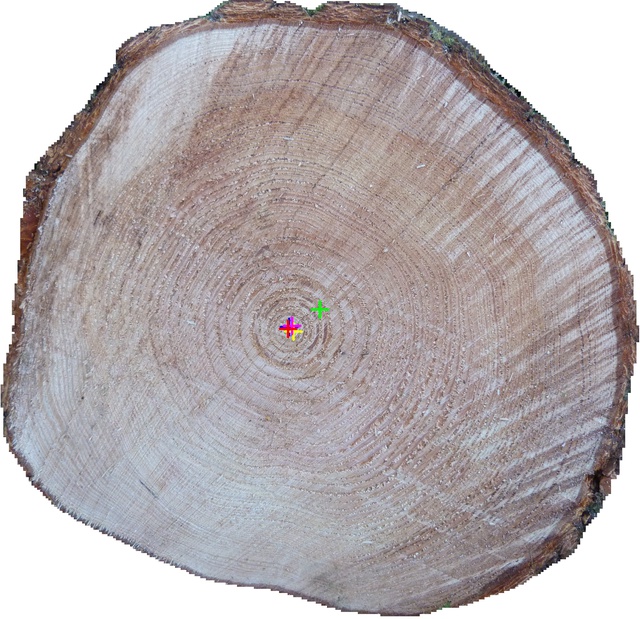}
    \label{fig:D03-S-4}
    \caption{D03-S-4}
    \end{centering}
    \end{subfigure}
    \hfill

    \begin{subfigure}{0.25\textwidth}
    \begin{centering}
    \includegraphics[width=\textwidth]{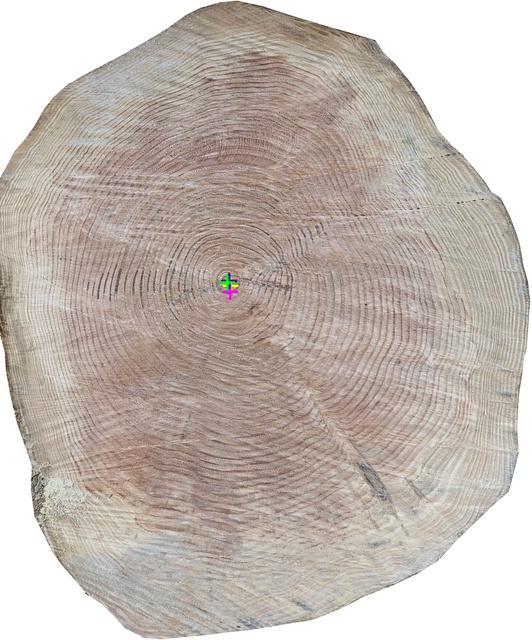}
    \label{fig:D13a}
    \caption{D13a}
    \end{centering}
    \end{subfigure}
    \hfill
    \begin{subfigure}{0.25\textwidth}
    \begin{centering}
    \includegraphics[width=\textwidth]{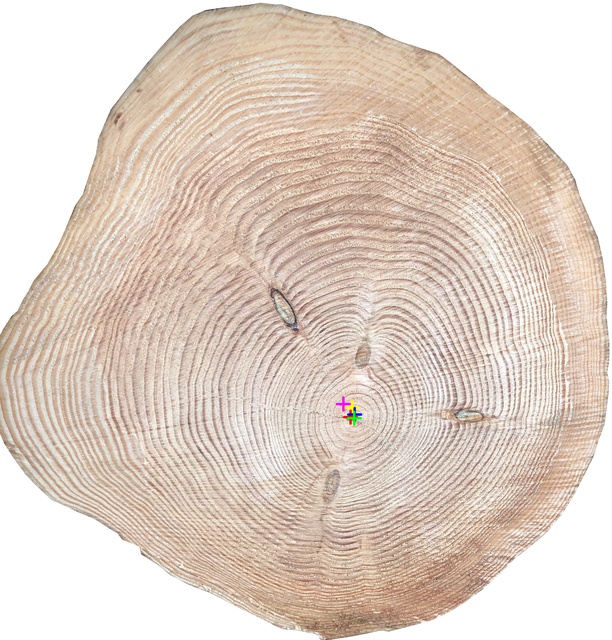}
    \label{fig:C05a}
    \caption{C05a}
    \end{centering}
    \end{subfigure}
    \hfill
    \begin{subfigure}{0.25\textwidth}
    \begin{centering}
    \includegraphics[width=\textwidth]{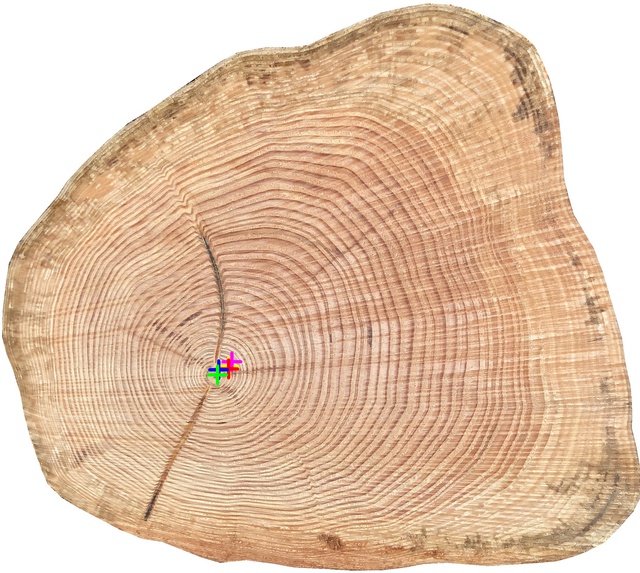}
    \label{fig:B07a}
    \caption{B07a}
    \end{centering}
    \end{subfigure}
    \hfill
    \begin{subfigure}{0.25\textwidth}
    \begin{centering}
    \includegraphics[width=\textwidth]{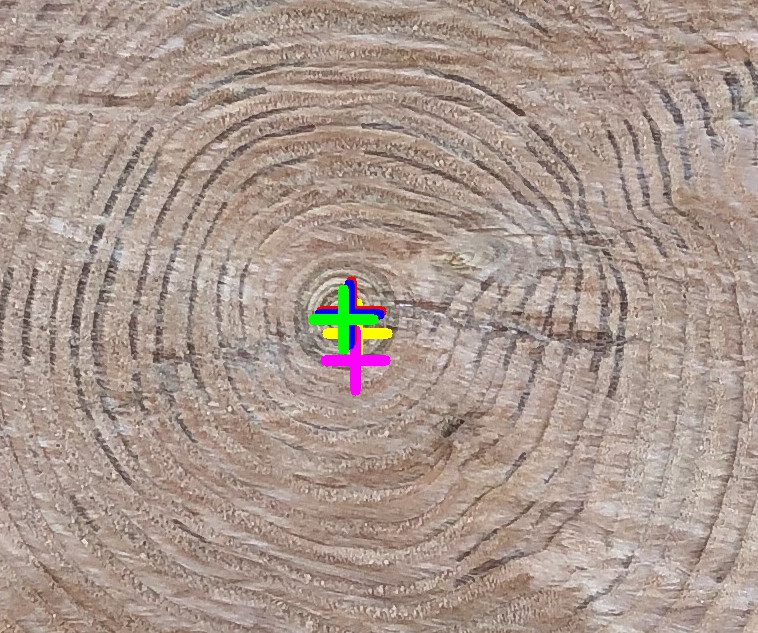}
    \label{fig:D13a_z}
    \caption{Zoom In D13a}
    \end{centering}
    \end{subfigure}
    \hfill
    \begin{subfigure}{0.25\textwidth}
    \begin{centering}
    \includegraphics[width=\textwidth]{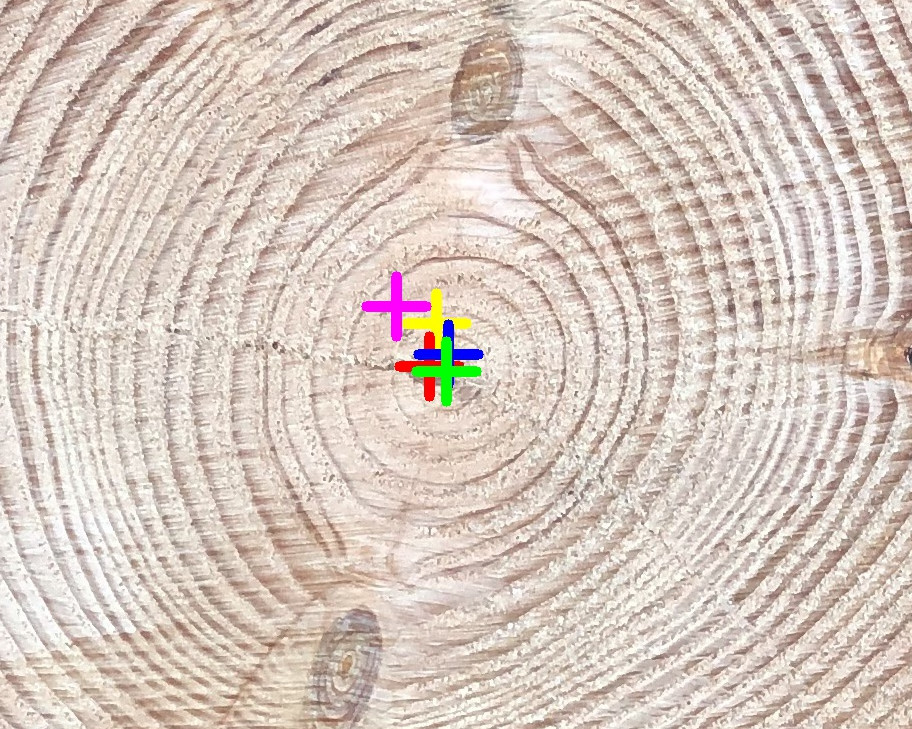}
    \label{fig:C05a_z}
    \caption{Zoom In C05a}
    \end{centering}
    \end{subfigure}
    \hfill
    \begin{subfigure}{0.25\textwidth}
    \begin{centering}
    \includegraphics[width=\textwidth]{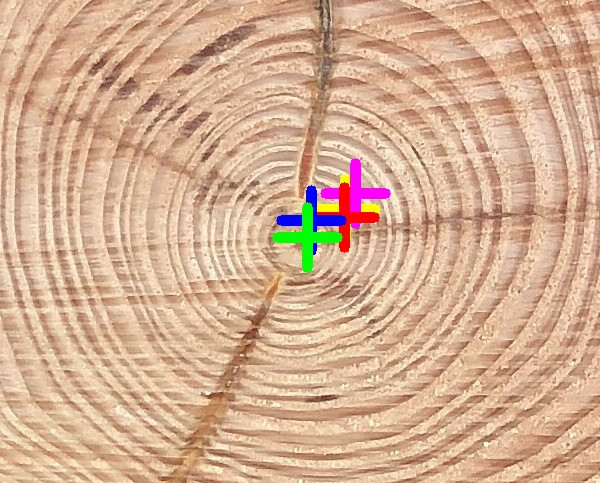}
    \label{fig:B07a_z}
    \caption{Zoom In B07a}
    \end{centering}
    \end{subfigure}
    \hfill
   \caption{Results over difficult cases, e.g fungus presence, no pith eccentricity (uncentred pith position) in  Forest, Logyard and Disc collections. Purple, LFSA; Red, ACO;  Blue, APD; Yellow, APD-PCL and Green, APD-DL}
   \label{fig:discussion_worst_cases_2}
\end{centering}
\end{figure}
 
\Cref{fig:discussion_worst_cases} illustrates how all the presented methods perform over the difficult cases in the UruDendro2 collection. \Cref{fig:discussion_worst_cases}.a to \ref{fig:discussion_worst_cases}.c illustrates cases where the ACO method (red marker) performs poorly. In one case, the method even predicts outside the disc region (\Cref{fig:discussion_worst_cases}.a). In this cases, the APD, APD-PCL and APD-DL (blue, yellow, and green markers, respectively) perform similarly, all of them close enough to the ground truth pith (the prediction is within the first tree ring region). On the other hand, the LFSA method (purple marker) performs slightly worse. 

\Cref{fig:discussion_worst_cases}.d to \ref{fig:discussion_worst_cases}.f show discs with a strong fungus presence around the pith position. In this case, the APD-PCL (yellow marker) performs considerably better than the other methods (in all cases, the prediction is within the first tree ring region).  When there is no tree ring information, the tree ring-based methods  (LFSA, ACO and APD) fail and do not converge near the pith position as illustrated in \Cref{fig:discussion_worst_cases}.e and \ref{fig:discussion_worst_cases}.f. This is the critical scenario for the ACO and APD methods. In the last stages, they search for the local minimum in a region centered around the previous pith prediction. The ring information will be very limited if the region is too small, which produces a method divergence. In the other hand, the APD-PCL method integrates radial structural insights derived from cracks and fungus in addition to the ring structure. Thereby enhancing its performance significantly in this particular scenario. Finally, the APD-DL method (green marker) also performs poorly in this scenario.

\Cref{fig:discussion_worst_cases_2} illustrates cases of off-center pith in Forest, Logyard and Disc collections and the presence of (strong) saw marks.  As seen in  \Cref{fig:discussion_worst_cases_2}.a to \ref{fig:discussion_worst_cases_2}.c, the APD-DL method performs considerably worse than the other methods. These disks belong to the Forest and Logyard collections. Both collections have 89 images, a small data size for training a deep learning method. As shown in the article \textit{Results} section, both APD and ACO outperform the APD-DL method on this dataset. 

\Cref{fig:discussion_worst_cases_2}.d to \ref{fig:discussion_worst_cases_2}.f, show images from the Discs collection in which the pith is strongly off-centered. A zoom-in is shown for each disc in \Cref{fig:discussion_worst_cases_2}.g to \ref{fig:discussion_worst_cases_2}.i. On this scenario, the LFSA method (purple marker) is less precise (in all the cases, it is further from the ground truth pith), whereas  APD and APD-DL perform similarly.

\section{Acknowledgments}
The experiments presented in this paper used ClusterUY \cite{clusterUy}  (site: https://cluster.uy). We had useful conversations with J. M. Morel and J. Di Martino.

\bibliographystyle{plainnat}
\bibliography{AutomaticWoodPithDetector}

\end{document}